\documentclass[10pt,twocolumn,letterpaper]{article}

\usepackage[pagenumbers]{cvpr} % To force page numbers, e.g. for an arXiv version

\usepackage[table]{xcolor}
\usepackage{amsmath}
\usepackage{amssymb}
\usepackage{mathtools}
\usepackage{amsthm}
\usepackage{caption}
\usepackage{subcaption}
\usepackage{multirow}
\usepackage{graphicx}
\usepackage{enumitem}
\usepackage{nicematrix}

\usepackage{algorithm}
\usepackage[noend]{algpseudocode}

\usepackage{svg}
\usepackage[export]{adjustbox}
\usepackage{etoolbox}
\usepackage{wrapfig}
\usepackage{changepage}

\definecolor{rowhighlight}{gray}{0.9}

\newcommand{\ours}{\texttt{SceneTok}\xspace}
\newcommand{\ourscene}{\texttt{SceneGen}\xspace}
\newcommand{\thickBox}{%
  \setlength{\fboxsep}{0pt}%
  \setlength{\fboxrule}{1.5pt}%
  \fbox{\rule{0pt}{0.5em}\rule{0.5em}{0pt}}%
}

\definecolor{tabfirst}{rgb}{1, 0.7, 0.7} % red
\definecolor{tabsecond}{rgb}{1, 0.85, 0.7} % orange
\definecolor{tabthird}{rgb}{1, 1, 0.7} % yellow
\definecolor{ours}{RGB}{0,176,255}
\definecolor{orange}{RGB}{255,192,0}
\definecolor{tabnone}{RGB}{128,128,128}
\definecolor{purple}{RGB}{112,48,160}

\newcommand{\baseline}[1]{\textcolor{black!60}{#1}}

\usepackage[table]{xcolor}
\usepackage{multirow}
\usepackage{hhline}
\usepackage{pifont}% http://ctan.org/pkg/pifont
\usepackage{appendix}

\newcommand{\xmark}{\ding{55}}%
\newcommand{\cmark}{\ding{51}}%

\definecolor{cvprblue}{rgb}{0.21,0.49,0.74}
\usepackage[pagebackref,breaklinks,colorlinks,allcolors=cvprblue]{hyperref}

\def\confName{CVPR}
\def\confYear{2026}

\title{\ours: A Compressed, Diffusable Token Space for 3D Scenes}

\author{Mohammad Asim \quad Christopher Wewer$^\dagger$
 \quad Jan Eric Lenssen\\
{\normalsize Max Planck Institute for Informatics, Saarland Informatics Campus}\\
{\tt\small \{masim, cwewer, jlenssen\}@mpi-inf.mpg.de} \\
{\small $^{\dagger}$ Corresponding author}
}

\begin{document}

\twocolumn[{%
\renewcommand\twocolumn[1][]{#1}%

\maketitle
\begin{center}
\centering
\captionsetup{type=figure}
\vspace{-0.4cm}
\includegraphics[width=1\textwidth]{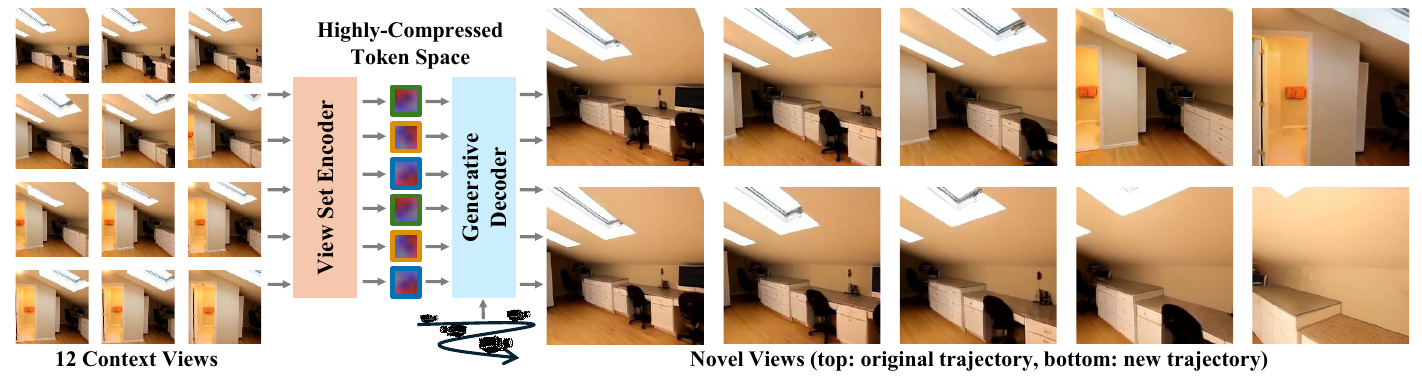}
\vspace{-0.7cm}
\captionof{figure}{\textbf{Setup Overview.} We introduce \ours, a tokenizer that encodes view sets into an unstructured, highly-compressed set of tokens, which can be efficiently rendered (32 images per second) from novel trajectories with a light-weight generative decoder. The token space is diffusable and allows latent generation of scenes in 5 seconds.}
\label{fig:teaser}
% \vspace{-0.1cm}
\end{center}
}]

\begin{abstract}
\vspace{-0.7cm}

We present \emph{\ours}, a novel tokenizer for encoding view sets of scenes into a compressed and diffusable set of unstructured tokens. Existing approaches for 3D scene representation and generation commonly use 3D data structures or view-aligned fields. In contrast, we introduce the first method that encodes scene information into a small set of permutation-invariant tokens that is disentangled from the spatial grid. The scene tokens are predicted by a multi-view tokenizer given many context views and rendered into novel views by employing a light-weight rectified flow decoder. We show that the compression is 1-3 orders of magnitude stronger than for other representations while still reaching state-of-the-art reconstruction quality. Further, our representation can be rendered from novel trajectories, including ones deviating from the input trajectory, and we show that the decoder gracefully handles uncertainty. Finally, the highly-compressed set of unstructured latent scene tokens enables simple and efficient scene generation in 5 seconds, achieving a much better quality-speed trade-off than previous paradigms. Code is available online: \url{geometric-rl.mpi-inf.mpg.de/scenetok/}

\vspace{-0.2cm}
\end{abstract}    

\vspace{-0.3cm}
\section{Introduction}
\label{sec:intro}

Answering the question of how to best represent 3D scenes in the era of large-scale, multi-modal, generative models is one of the most pressing research milestones for several applications in computer vision and graphics, such as reconstruction~\cite{mildenhall2020nerf, kerbl3Dgaussians, yu2021pixelnerf, pixelsplat, latentsplat, mvsplat, gslrm, longlrm, lvsm2025, rayzer, mitchel2025true}, open-world scene understanding~\cite{Zhang_2021_CVPR, Nie_2020_CVPR, huang2018cooperative}, reasoning~\cite{wewer25srm, ma2025spatialreasoner, Chen_2024_CVPR, cheng2024spatialrgpt}, or generation of 3D worlds~\cite{bolt3d, cat3d, dfm, chan2023genvs, dreamgaussians, yi2023gaussiandreamer, dfot, seva, zhao2024genxd}. Such representations come with several requirements: (1) being trainable from a data source that is available in vast quantities, like videos, (2) being simple in structure to expose an easy interface to large-scale, potentially multi-modal generative models, (3) being highly compressive to handle the large-scale nature of 3D scenes. In this work, we make an important step towards this goal by introducing \ours, an autoencoder that encodes 3D scenes from videos into a set of unstructured, highly compressed tokens, which can be rendered from novel views and can be processed by generative diffusion models to perform conditional generation in latent space.

Existing paradigms for 3D scene generation often use an underlying 3D data structure ~\cite{dfm, chan2023genvs, dreamgaussians, yi2023gaussiandreamer} or posed multi-view images and videos~\cite{cat3d, dfot, seva, bolt3d, yu2024viewcrafter}. The former has several limitations. Due to the lack of 3D data and the cubic scale of 3D-structured representations, training of large foundation models is prohibitively expensive or impossible. The latter enables generative modeling by learning from large-scale video data. However, these models are typically large and require specialized sampling strategies such as history-guided and autoregressive generation ~\cite{dfot, diffforce} or anchored generation~\cite{cat3d, dfot, seva, bolt3d, yu2024viewcrafter} to produce large and consistent scenes; therefore, they necessitate immense computing effort due to the simultaneous generation of coarse structures and rendered details. Recent works in generalizable 3D reconstruction~\cite{srn, lvsm2025, rust, rayzer} encode images into a simplified latent space that is disentangled from the spatial grid.
However, they have been shown to perform only view interpolation~\cite{mitchel2025true}, i.e., they cannot be rendered from novel views, and due to high dimensionality, these methods do not enable generation or other downstream tasks.

In contrast, \ours employs a simple 2-stage approach. In the first stage, we train an autoencoder to compress a given set of context views into an unstructured set of continuous tokens. The representation can subsequently be decoded by a lightweight, diffusion-based renderer, sampling multiple novel views in less than a second, and accounting for the remaining uncertainty in the scene representation. To show the usefulness of our latent space, we introduce a second stage, which consists of a diffusion model trained on the scene token space from stage one, enabling view-conditioned generation. The cascaded approach changes the generative paradigm from view-space to a compressed latent-space and \emph{decouples view rendering from scene generation}. This allows allocating more resources to the generative model, i.e., scaling it up without affecting the rendering speed. Furthermore, the compressed space makes it highly efficient to perform generation.

We evaluate both stages independently and show that \ours achieves state-of-the-art reconstruction quality, can render novel trajectories, and can gracefully resolve uncertainty in the scene representation. Decoding is efficient, allowing the rendering of 32 novel views in 1 second on an Nvidia RTX 4090. Meanwhile, our latent diffusion model on the obtained representation can effectively perform conditional 3D scene generation in 5 seconds, which is orders of magnitude faster than other scene generation paradigms.
In summary, we propose:
\begin{itemize}
    \item a novel paradigm for scene generation that decouples \emph{view rendering} from \emph{generation}. 
    \item a compressed, unstructured token representation for scenes that enables fast \emph{rendering} from novel views using a scene autoencoder (\ours) that encodes and decodes a set of posed views of a scene to and from a compressed token set with a generative renderer.

    \item a diffusion transformer model (\ourscene) that performs \emph{generation} in the latent space very efficiently.

\end{itemize}

\section{Related Work}
\label{sec:related_works}

\ours is a 3D scene tokenizer for efficient generation. It can be seen at the intersection of vision tokenizers for latent image and video generation (cf. Sec.~\ref{sec:vision_tokenizers}), 3D scene reconstruction (c.f. Sec.~\ref{sec:generalizable_3d}), and generation (c.f. Sec.~\ref{sec:3d_scene_generation}).

\subsection{Vision Tokenizers for Latent Generation} \label{sec:vision_tokenizers}
Scalable and efficient image and video generation with autoregressive transformers~\cite{imgtransformer} or diffusion models~\cite{ddpm} relies on autoencoders, also known as vision tokenizers, that compress data into compact latent spaces.

\vspace{-0.4cm}
\paragraph{Spatial Autoencoders.} Pioneering works VQGAN~\cite{esser2020taming} and Latent Diffusion~\cite{ldm} leverage 2D convolutional autoencoders for spatial perceptual compression of images into discrete codebook indices or continuous latents, respectively. In both cases, the latent space maintains the 2D image structure but in a lower resolution.
This paradigm has been successfully extended to 3D convolutional tokenizers for videos~\cite{cvvae, cogx, opensora2}.
Moreover, recent follow-up works improve this family of autoencoders in terms of stronger compression~\cite{dcae} and "diffusability" of the latent space~\cite{vavae}.

\vspace{-0.4cm}
\paragraph{1D Tokenizers.}
Another new line of research proposes to encode images~\cite{titok,flextok,semanticist} and videos~\cite{larp} into unstructured token sets or 1D sequences.
These are directly compatible with autoregressive generation, whereas 2D or 3D patch tokens for images and videos lack an intuitive order.

With \ours, we propose the first tokenizer that compresses 3D scenes into an unstructured token set. While grid-like representations from spatial autoencoders are computationally feasible for images and videos, the cubic cost of voxel grids severely limits scalability.
%Alternative representations have been explored in the context of generalizable 3D scene reconstruction.

\begin{figure*}[t]
    \centering
    \begin{subfigure}[t]{.56\textwidth}
        \centering
        \includegraphics[width=\textwidth]{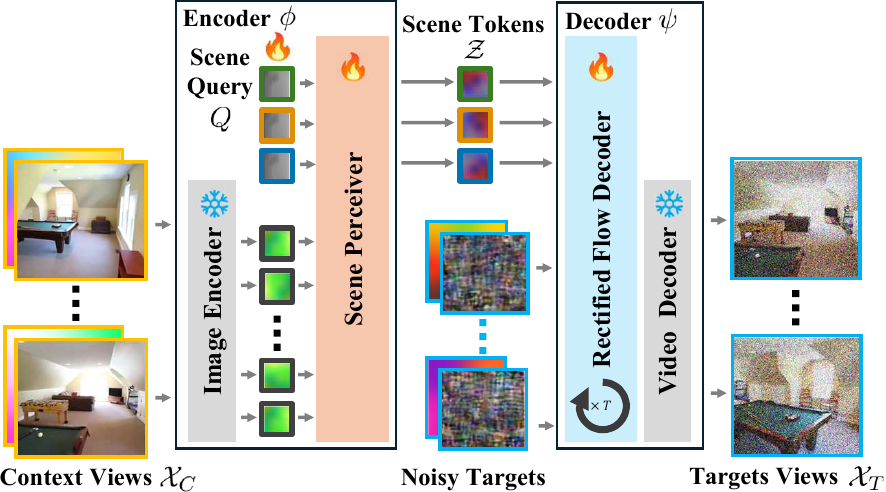}
        \caption{\ours Autoencoder}
        \label{fig:method-ae}
    \end{subfigure}
    % \hspace*{\fill}%
    \begin{subfigure}[t]{.37\textwidth}
        \centering
        \includegraphics[width=\textwidth]{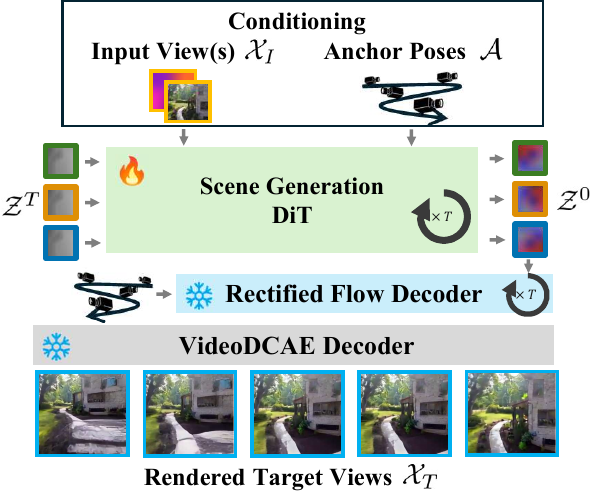}
        \caption{Latent Scene Token Generation}
        \label{fig:method-gen}
    \end{subfigure}
    \vspace{-0.2cm}
    \caption{\textbf{Method Overview.} \textbf{(a)} The \ours autoencoder encodes view sets into a set of compressed, unstructured scene tokens by chaining a VA-VAE image compressor and a perceiver module. The tokens can be rendered from novel views with a generative decoder based on rectified flows. \textbf{(b)} A latent diffusion transformer can perform scene generation by generating compressed scene tokens. Scene generation can be conditioned on a single or a few images and a set of anchor poses, defining the spatial scene extent. \label{fig:method} 
    }
    \vspace{-0.4cm}
\end{figure*}

\subsection{Feed-Forward 3D Scene Reconstruction}
\label{sec:generalizable_3d}
A feed-forward 3D reconstruction pipeline can be seen as an autoencoder that encodes a set of input views and reconstructs a set of target views. An important design choice is the intermediate representation encoding the 3D scene.

\vspace{-0.4cm}
\paragraph{Explicit Representation.} Several works encode input views into 3D Gaussians \cite{pixelsplat,mvsplat,latentsplat,depthsplat,longlrm,gslrm,mvsplat360} and NeRFs~\cite{pixelnerf, ibrnet, mvsnerf, geonerf} by directly predicting them in a feedforward manner. However, both NeRFs and 3D Gaussians have large representation sizes that are high-dimensional in nature and thus follow a complex distribution, making them ill-suited for efficient and scalable generation.

\vspace{-0.4cm}
\paragraph{Latent Representation.} Closer to our method are LVSM~\cite{lvsm2025} and RayZer~\cite{rayzer}, which do not impose any 3D inductive bias but predict novel views directly using large-scale transformer architectures. However, both leverage more than 3K high-dimensional ($\geq512$) tokens which makes latent diffusion on their representations infeasible. Moreover, in addition to information leakage in RayZer~\cite{rayzer}, concurrent work shows that it lacks transferability~\cite{mitchel2025true}, i.e., the ability to synthesize novel views that deviate from the original input camera trajectory.

In contrast, \ours is the first 3D scene tokenizer that produces a highly compressed representation, well-suited for generation via latent diffusion and allows for transferable novel view synthesis.

\subsection{Generative Models for 3D Scenes} \label{sec:3d_scene_generation}
 Scene generation has largely evolved from optimization-based 3D to feed-forward view-space generation, with the availability of large-scale video and multi-view datasets facilitating state-of-the-art generation performance.
\vspace{-0.3cm}
\paragraph{Generation in 3D-Space.} Several works~\cite{nichol2022pointegenerating3dpoint, jun2023shapegeneratingconditional3d, dfm, liu2023zero1to3, wonder3d, roessle2024l3dg} allow direct generation of 3D representations, either via Score-Distillation-Sampling (SDS) or by baking 3D representations into the architecture. This results in 3D consistent generation~\cite{met3r}; however, they are not scalable with the size of the scene and are mostly limited to object-centric scenes. Overall, the training and inference of such models require immense computational effort and resources, while large-scale 3D datasets remain very limited.  
\vspace{-0.3cm}
\paragraph{Generation in View-Space.} Conditional multi-view~\cite{cat3d, seva} and video~\cite{yu2024viewcrafter, dfot} diffusion models generate target views directly conditioned on input view(s). While they can leverage the availability of large-scale datasets for pre-training, rendering novel views still requires expensive generation steps, as the models are typically large to ensure good generation quality. However, the entanglement of view rendering with generation causes redundancies, e.g., when revisiting the same views that were previously generated, resulting in a waste of computing resources. 

In this work, we leverage the benefits of both worlds: (1) the disentanglement of \emph{rendering} and \emph{generation} , and (2) utilizing large-scale multi-view and video data. Additionally, we can allocate more capacity towards generation and enable fast rendering with a lightweight decoder. 

\section{SceneTok}
\label{sec:method}
The core of our method is an autoencoder (c.f. Fig.~\ref{fig:method-ae}), which takes a set of $N$ context images $\mathcal{X}_C = \{\mathbf{x}_i\}_{i=1}^{N}$ as observations of a 3D scene with camera poses $\mathcal{P}_C = \{\mathbf{p}_i\}_{i=1}^{N}$ and encodes them via the encoder $\phi$, i.e.
\begin{equation}
    \mathcal{Z} := \{\mathbf{z}_i\}_{i=1}^K = \phi(\mathcal{X}_C, \mathcal{P}_C)\textnormal{,}
\end{equation}
where $\{\mathbf{z}_i \in \mathbb{R}^d\}_{i=1}^K$ is a set of continuous tokens. Then, a decoder $\psi$ takes tokens $\mathcal{Z}$ and a novel camera trajectory $\mathcal{P}_T:=\{\mathbf{p}_i\}_{i=1}^M$ and renders it, i.e.
\begin{equation}
    \mathcal{X}_T = \psi(\mathcal{Z}, \mathcal{P}_T) \textnormal{,}
\end{equation}
where $\mathcal{X}_T:=(\mathbf{x}_i)_{i=1}^M$ is a sequence of $M$ novel views for the given camera poses, which can be supervised (during autoencoder training), or serve as output during inference. In the second stage, we formulate a generative model on the token set $\mathcal{Z}$ (c.f. Fig.~\ref{fig:method-gen}). We will introduce the encoder in Sec.~\ref{sec:encoder}, the decoder in Sec.~\ref{sec:decoder}, the training in Sec.~\ref{sec:training}, and the latent generative model  in Sec.~\ref{sec:generator}.

\subsection{Encoding the Scene}
\label{sec:encoder}
\ours predicts an unstructured set of tokens $\mathcal{Z} := \{\mathbf{z}_i\}_{i=1}^K$ as the output of the encoder $\phi$, which forms our latent scene representation. During encoding, we first compress each image $\mathbf{x}_i\in\mathbb{R}^{H\times{W}\times{3}}$ into a latent feature map \mbox{$\mathbf{f}_i\in\mathbb{R}^{\frac{H}{16}\times\frac{W}{16}\times32}$} using the VA-VAE encoder~\cite{vavae} with 16x spatial compression. 
These features from all views \mbox{$\mathcal{F}_C:=\{\mathbf{f}_i\}_{i=1}^{N}$} are fed as $1\times1$ latent patch tokens into a dedicated branch of the following scene perceiver. Inside each block, camera poses $\mathcal{P}_C$ are transformed into ray maps \mbox{$\mathcal{R}_C=\{\mathbf{r}_i\in\mathbb{R}^6\}^{N}_{i=1}$} with $\mathbf{r}_i=[\mathbf{o}_i, \mathbf{d}_i]$ for ray origins $\mathbf{o}_i$ and directions $\mathbf{d}_i$, which are used to modulate the corresponding patch tokens via AdaLN~\cite{dit} before a multi-view attention layer and an MLP.

\begin{figure}
    \centering
    \includegraphics[width=1.0\linewidth]{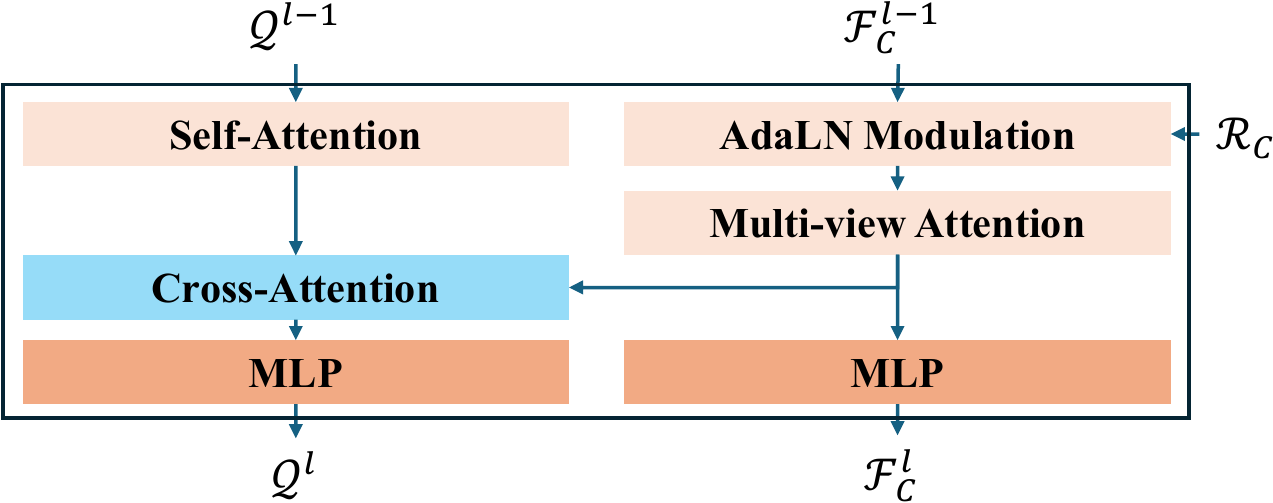}
    \vspace{-0.5cm}
    \caption{\textbf{A single scene perceiver block.} The scene perceiver module consists of $L$ blocks, alternating self-attention between scene queries $\mathbf{Q}$ and cross-attention to the multi-view encoder branch, which is AdaLN-modulated with ray embeddings.}
    \label{fig:encoder_block}
    \vspace{-0.4cm}
\end{figure}

A second branch of the perceiver, dedicated to the scene tokens, processes a set of directly optimized scene queries \mbox{$Q:=\{q_i\}_{i=1}^{K}$} with $q_i\in\mathbb{R}^{d_H}$ (c.f. Fig.~\ref{fig:encoder_block}). Each block of this branch consists of self-attention and cross-attention to the intermediate multi-view patch tokens of the first branch. Figure~\ref{fig:encoder_block} shows an individual encoder block with $\mathcal{F}_C^0=\mathcal{X}_C$ and $\mathcal{Q}^0=\mathcal{Q}$. We obtain the scene tokens $\mathcal{Z}$ after an additional projection to a lower dimension $d$. We provide hyperparameter and architectural details in Appendix Sec.~\ref{sec:training_architecture_add}.

\vspace{-0.3cm}
\paragraph{Choice of Positional Encoding.}
Each visual token from the context views is encoded using rotary positional encodings (RoPE)~\cite{rope}. We found that invariances defined by positional encodings strongly influence the scene representation. 3D RoPE~\cite{rope3d}, as is common in video encoders, gives the context views a temporal order perceivable by the encoder and leads to an unwanted temporal bias. Thus, we employ only 2D RoPE, which leads to an order invariant encoder and ensures that the scene tokens can be rendered from arbitrary trajectories later. More details and findings are provided in Appendix Sec.~\ref{sec:scene_token_analysis_add}.
 
\subsection{Decoding the Scene}
\label{sec:decoder}
A scene representation can lack information about details that are necessary to render certain novel views in sufficient quality. In \ours, there are two common sources of uncertainty: (1) information that is already missing from the input views, or (2) high-frequency details that are lost due to compression. Both lead to varying uncertainty when rendering novel views, depending on the specific view. We find that it is crucial to resolve this uncertainty via sampling, which requires a generative renderer $\psi$ to decode the representation $\mathcal{Z}$ by sampling from a conditional distribution $p_\Psi(\mathbf{x}|\mathcal{Z})$. Note that the renderer can also adapt to varying uncertainty based on location. For regions that are clearly defined by the scene tokens, it can learn to sample from a narrow distribution while falling back to pure generation in cases of high uncertainty. In Figure~\ref{fig:variance_vs_mask}, we show that there is a correlation between the variance of rendered outputs and the information in the tokens. 
\begin{table*}
    \centering
    \resizebox{\linewidth}{!}{
     \begin{NiceTabular}{l|cccccc|cccccc} 
     \toprule
     & \multicolumn{6}{c}{12 Context Views} 
     & \multicolumn{6}{c}{5 Context Views} \\ 
    & Repr. Size (\#Floats) &PSNR $\uparrow$ & LPIPS $\downarrow$ & SSIM $\uparrow$ & rFVD $\downarrow$ & rFID $\downarrow$
    & Repr. Size (\#Floats) & PSNR $\uparrow$ & LPIPS $\downarrow$ & SSIM $\uparrow$ & rFVD $\downarrow$ & rFID $\downarrow$\\ 
        \midrule
        \baseline{\emph{Explicit Representation}} \\

        \baseline{MVSplat}~\cite{mvsplat} & \baseline{46.40M} & \baseline{20.13} &  \baseline{0.255} &  \baseline{0.755} & \baseline{354.46} & \baseline{29.89}& \baseline{19.33M} &  \baseline{22.82} &  \baseline{0.152} &  \baseline{0.856} & \baseline{144.91} & \baseline{15.85}\\
        \baseline{MVSplat360}~\cite{mvsplat360}    & \baseline{74.71M}      & \baseline{21.09} &  \baseline{0.247} &  \baseline{0.769} & \baseline{294.82} &  \baseline{25.64}&\baseline{31.12M} &\baseline{23.88} & \baseline{0.174} & \baseline{0.828} & \baseline{136.52} & \baseline{18.15}\\

        \baseline{DepthSplat}~\cite{depthsplat} & \baseline{46.40M}    & \baseline{21.55} &  \baseline{0.202} &  \baseline{0.809} & \baseline{204.52} & \baseline{21.35}  &  \baseline{19.33M} & \baseline{24.45} &  \baseline{0.124} &  \baseline{0.884} & \baseline{109.74} & \baseline{12.89} \\
        
        \midrule
        \emph{Latent Representation}\\
        LVSM~\cite{lvsm2025} & \cellcolor{tabsecond}1.57M & \cellcolor{tabsecond}21.25 &  \cellcolor{tabsecond}0.262 &  \cellcolor{tabsecond}0.686 & \cellcolor{tabsecond}211.66 & \cellcolor{tabsecond}26.40 & \cellcolor{tabsecond}1.57M & \cellcolor{tabsecond}25.74 &  \cellcolor{tabsecond}0.140 &  \cellcolor{tabfirst}\textbf{0.831} & \cellcolor{tabsecond}111.14 & \cellcolor{tabsecond}13.37 \\
        \textbf{SceneTok}     & \cellcolor{tabfirst}\textbf{32.76K}    &  \cellcolor{tabfirst}\textbf{23.99} &  \cellcolor{tabfirst}\textbf{0.159} &   \cellcolor{tabfirst}\textbf{0.783} &  \cellcolor{tabfirst}\textbf{79.80} &  \cellcolor{tabfirst}\textbf{11.12}  & \cellcolor{tabfirst}\textbf{32.76K}&  \cellcolor{tabfirst}\textbf{25.97} &  \cellcolor{tabfirst}\textbf{0.133} &  \cellcolor{tabsecond}0.817 &  \cellcolor{tabfirst}\textbf{76.24} &  \cellcolor{tabfirst}\textbf{11.26}\\
     \bottomrule
    \end{NiceTabular}
    }
   \vspace{-0.2cm}

    \caption{\textbf{Quantitative comparison of NVS quality on RE10K.} \ours achieves superior performance compared to previous works achieving significant improvements in terms of rFVD and rFID alongside with PSNR and LPIPS while having a significantly smaller representation size.   
    }
    \label{tab:nvs_re10k}
    \vspace{-0.4cm}
\end{table*}
\vspace{-0.3cm}
\paragraph{Decoder Architecture.} We implement the rendering function $\psi$ as a two-step procedure. During inference, we first employ a rectified flow generator $\Psi$ that iteratively denoises latent image patches for novel trajectories, according to a reverse process from $\mathbf{x}^1$ to $\mathbf{x}^0$. One step of the denoising process computes
\begin{equation}
    \mathbf{x}^{t-\Delta t}=\mathbf{x}^t - \Delta t\Psi(\mathbf{x}^t, \mathcal{R}, \mathcal{Z}, t) \textnormal{, }
\end{equation}
conditioned on $t$, ray maps $\mathcal{R}$ of a novel trajectory, and scene tokens $\mathcal{Z}$, where $\mathbf{x}^1$ is initialized from a standard Gaussian distribution and $\Delta t$ is the step size. We use flow-parameterization for rectified flow, predicting a vector field (c.f. Sec.~\ref{sec:training}). The resulting $\mathbf{x}^0$ are decoded into pixel space by the VideoDCAE~\cite{opensora2} decoder from Open-Sora 2.0.

Our diffusion network is a diffusion transformer model and closely resembles that of LightningDiT~\cite{vavae}. We remove class conditioning and instead condition each latent token on its corresponding ray map token by merging it with the timestep embedding~\cite{dfot} and applying AdaLN, as in Sec.~\ref{sec:encoder}. In each block, we condition on the scene tokens $\mathcal{Z}$ via a cross-attention layer. We employ latent-space diffusion over pixel space due to its efficiency, which is essential for the fast rendering of scene tokens via conditional generation with a video diffusion model. We provide further discussions and  technical details in Appendix Sec.~\ref{sec:training_architecture_add}.

\subsection{Training}
\label{sec:training}
We keep the VA-VAE encoder and VideoDCAE decoder frozen while training the rest of the architecture end-to-end with the rectified flow matching objective~\cite{lipman2023flow, liu2023flow} in latent space. Specifically, for each output view, the output of the decoder is a vector field, i.e.
\begin{equation}
    \mathbf{v}^t=\Psi(\mathbf{x}^t, \mathcal{R}, \mathcal{Z}, t) \textnormal{.}
\end{equation}
Then, the objective function is a simple MSE between the ground truth flow $\mathbf{x}^1 - \mathbf{x}^0$ and the predicted flow:
\begin{equation}
\label{eq:rf}
%p_{\psi, t}(\mathbf{x}^t | \mathcal{Z}, \mathcal{R}, \mathbf{x}^1), p(\mathbf{x}^1)}
    \mathcal{L}(\psi) = \mathbb{E}_{t, \mathbf{x}^1, \mathbf{x}^0} ||(\mathbf{x}^1 - \mathbf{x}^0) - \mathbf{v}^t||_2^2 \textnormal{,}
\end{equation}
where $\mathbf{x}^t = t\mathbf{x}^1 + (1-t)\mathbf{x}^0$, $\mathbf{x}^1\sim N(\mathbf{0}, \mathbf{I})$, and $\mathbf{x}^0$ are latent target images from the training set.
When training end-to-end (i.e., when $\mathcal{Z}$ is parameterized by the encoder $\phi$), we effectively learn to extract a representation $\mathcal{Z}$ from the observations $\mathcal{X}_C$ and $\mathcal{P}_C$ that maximizes the likelihood of rendering our target views. Meanwhile, the decoder can learn to adapt to varying conditioning strength from $\mathcal{Z}$. See Sec.~\ref{par:transferability} and~\ref{sec:analysis} for further analysis on the $\mathcal{Z}$. 

\subsection{Latent Scene Generation}
\label{sec:generator}
Once we trained the autoencoder, we can freeze both the encoder and decoder and train a simple diffusion transformer~\cite{yao2024fasterdit} to model the conditional distribution $p(\mathcal{Z}|\mathcal{X}_I,\mathcal{A})$ over the scene tokens $\mathcal{Z}$ with \mbox{$\mathcal{A}:=\{\mathbf{a}_i\in\mathbb{R}^6\}_{i=1}^N$} being camera anchors and $\mathcal{X}_I$ a single or a few conditioning image(s). A single anchor is defined as $\mathbf{a}_i=[\mathbf{o}_i,\mathbf{d}_i]$, where $\mathbf{o}_i, \mathbf{d}_i\in\mathbb{R}^3$ are the $i^{th}$ position and viewing direction of a camera in 3D space. We embed the ray map and take a single averaged embedding over all pixel to minimize the number of tokens. The complete set $\mathcal{A}$ defines the spatial extent and layout of the scene that is to be generated. Conditioning on such anchors allows us to control how much the scene tokens must span and helps define the camera trajectories for rendering novel views later. Because we center our scenes around the origin during encoding (c.f. Sec.~\ref{sec:encoder}), we can always set the first input image and the anchor to the scene origin and set the rotation to the identity. The subsequent camera poses for the anchors, input, and target views are provided relative to this origin. The scene generation transformer, which we denote as \ourscene in the experiments, is also trained using rectified flows (c.f. Eq.~\ref{eq:rf}) on a dataset of encoded scene tokens.  We provide training details in Sec.~\ref{sec:experimental_setup}.

\section{Experiments}
\label{sec:experiments}

\begin{figure*}[t]
    \centering
    \makebox[\linewidth][c]{
        \centering
        \subfloat[RealEstate10K (12 Context Views)]{%
        \includegraphics[width=0.495\linewidth]{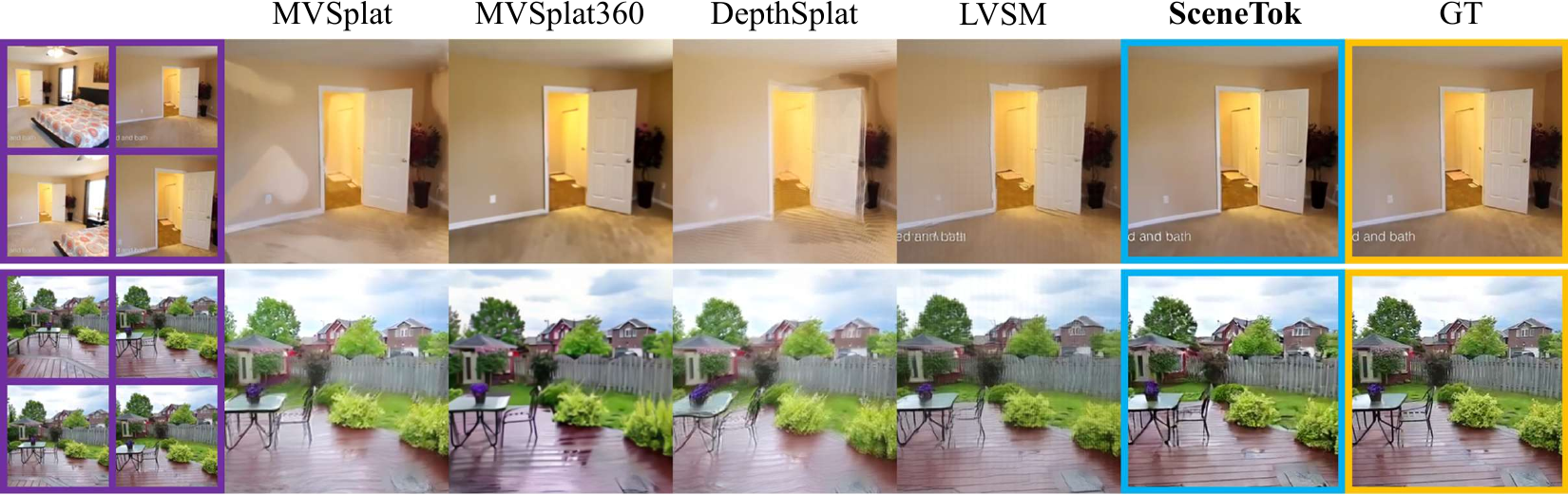}
        \includegraphics[width=0.495\linewidth]{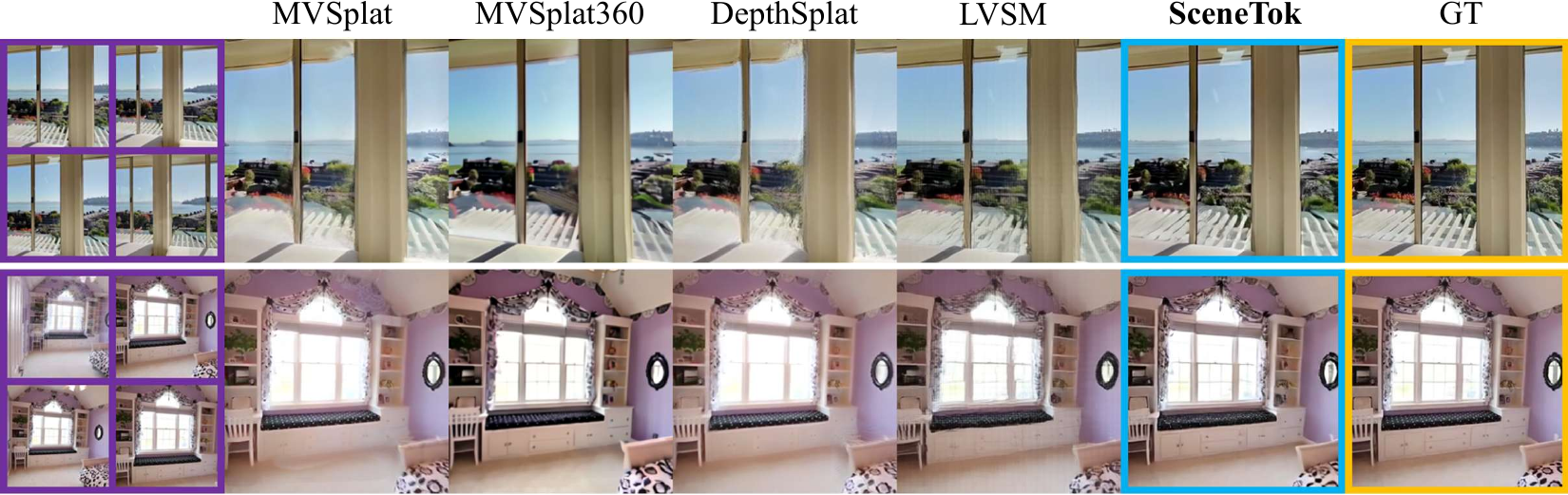}

        }
    }
  \makebox[\linewidth][c]{
      \centering

    \subfloat[Dl3DV-140 (16 Context Views)]{%
      \includegraphics[width=0.495\linewidth]{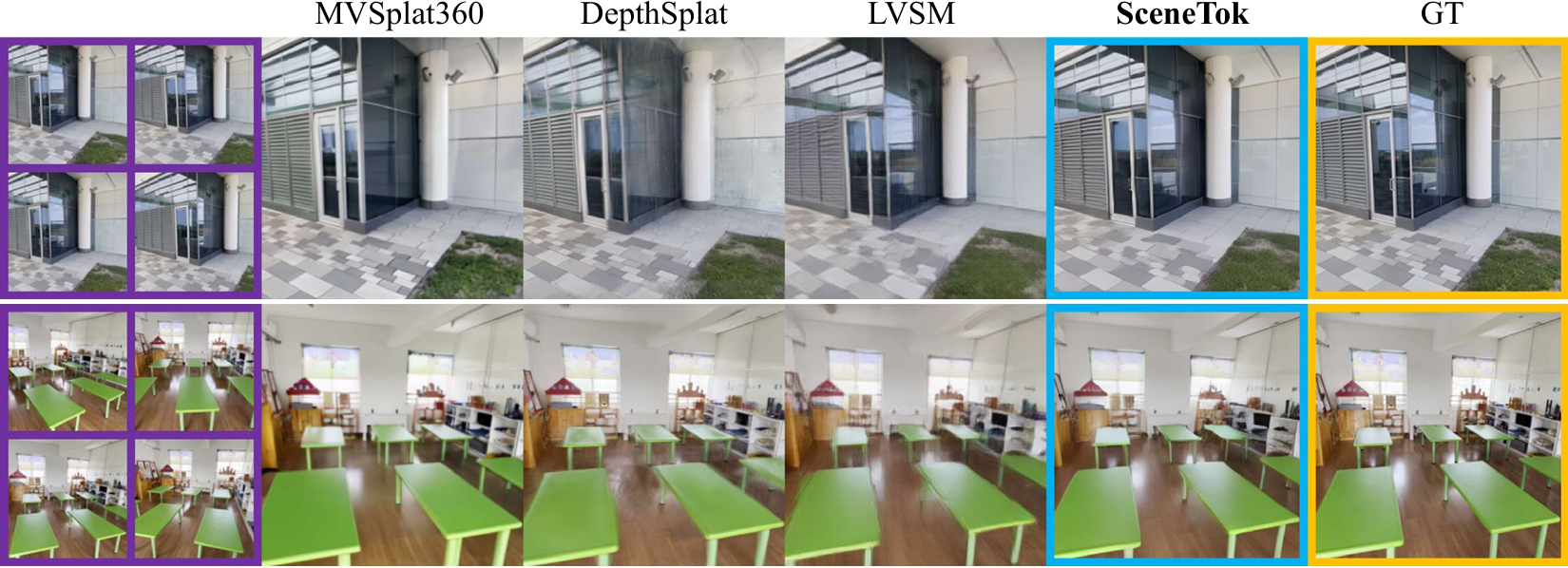}%
    }\hspace{0.001\linewidth}
    \subfloat[Zero-Shot ACID (12 Context Views)]{%
      \includegraphics[width=0.495\linewidth]{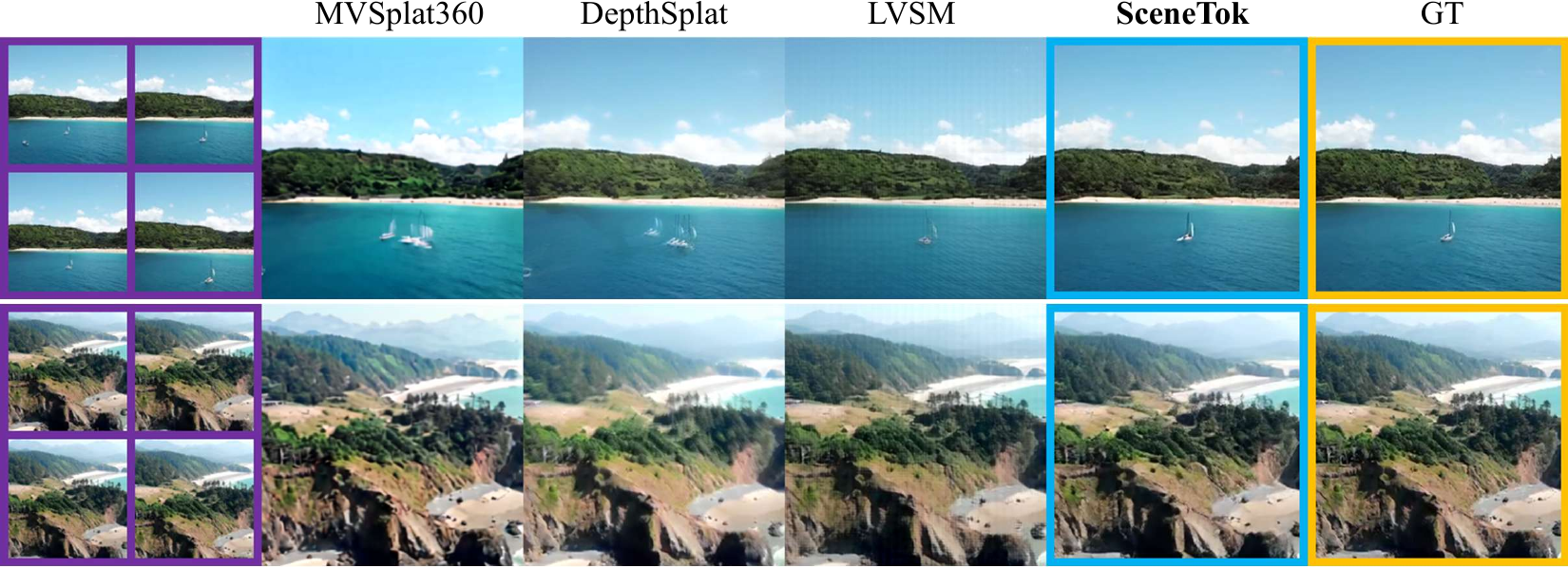}%
    }
    }

    \vspace{-0.2cm}
    \caption{\textbf{Qualitative NVS.} $($\textcolor{purple}{\thickBox}, \textcolor{ours}{\thickBox}, \textcolor{orange}{\thickBox}$)$ denote the context views (only four is shown), the target rendering of \ours and the ground-truth target view (only one is shown) respectively. While the baselines suffer from blur artifacts even without compression, \ours produces cleaner renderings and better details from a small set of scene tokens. More examples in Appendix Sec.~\ref{sec:qualitative_results_add}.
    }
    \label{fig:qualitative_nvs}
    \vspace{-0.2cm}
\end{figure*}

In this section, we devise a set of experiments that showcase the potential of our method. We choose baselines that solve novel view synthesis and scene generation with different paradigms and aim to answer the following questions:

\begin{itemize}
\itemindent=16pt 
\item[\textbf{Q1:}] How does the \ours autoencoder fare against prior works in terms of novel view synthesis quality?
\item[\textbf{Q2:}] Do the scene tokens satisfy the properties of a 3D representation, i.e., allow rendering from novel trajectories? 
\item[\textbf{Q3:}] Does the compressed token representation allow for efficient scene generation in latent space and how does it compare against multi-view and video generators?
\end{itemize}
We define our experimental setup in Sec.~\ref{sec:experimental_setup}, answer Q1 and Q2 in Sec.~\ref{sec:nvs}, and finally Q3 in Sec.~\ref{sec:scene_generation}.

\subsection{Experimental Setup}
\label{sec:experimental_setup}
We devise a series of experiments to evaluate the novel-view synthesis quality and transferability~\cite{mitchel2025true} of \ours on novel trajectories, as well as the single-view generation quality of \ourscene.  
\begin{table*}
    \centering
    \resizebox{\linewidth}{!}{
     \begin{NiceTabular}{l|cccccc|cccccc}
     \toprule
     & \multicolumn{6}{c}{DL3DV-140} 
     & \multicolumn{6}{c}{Zero-Shot ACID} \\ 
    & Repr. Size (\#Floats) & PSNR $\uparrow$ & LPIPS $\downarrow$ & SSIM $\uparrow$ & rFVD $\downarrow$ & rFID $\downarrow$ 
    & Repr. Size (\#Floats) & PSNR $\uparrow$ & LPIPS $\downarrow$ & SSIM $\uparrow$ & rFVD $\downarrow$ & rFID $\downarrow$\\ \midrule

\baseline{\emph{Requires Target Views}}\\
\baseline{RayZer}~\cite{rayzer} & \baseline{2.36M} & \baseline{26.89} & \baseline{0.154} & \baseline{0.834} & \baseline{87.41} & \baseline{14.75} & \cellcolor{tabnone}- & \cellcolor{tabnone}-& \cellcolor{tabnone}-& \cellcolor{tabnone}-& \cellcolor{tabnone}-& \cellcolor{tabnone}- \\
\midrule
\baseline{\emph{Explicit Representation}}\\

\baseline{MVSplat360}~\cite{mvsplat360}&\baseline{174.32M} %/ 99.61M 
& \baseline{18.54} & \baseline{0.278} & \baseline{0.629} & \baseline{314.28} & \baseline{21.95}  & \baseline{74.71M} & \baseline{21.75} & \baseline{0.280} & \baseline{0.701} & \baseline{294.86} & \baseline{43.24}\\
\baseline{DepthSplat}~\cite{depthsplat} &
\baseline{108.27M} %/ 61.89M  
& \baseline{21.86} & \baseline{0.192} & \baseline{0.769} & \baseline{423.91} & \baseline{20.90} & \baseline{46.40M} & \baseline{21.52} & \baseline{0.216} & \baseline{0.772} & \baseline{351.97} & \baseline{29.20} \\

\baseline{Long-LRM}~\cite{longlrm} &\baseline{492.99M} %/ 61.89M 
& \baseline{23.06} & \baseline{0.181} & \baseline{0.796} &\baseline{519.19} & \baseline{26.56}  & \cellcolor{tabnone}- & \cellcolor{tabnone}-& \cellcolor{tabnone}-& \cellcolor{tabnone}-& \cellcolor{tabnone}- & \cellcolor{tabnone}-\\

\midrule

\emph{Latent Representation}\\

LVSM~\cite{lvsm2025} & \cellcolor{tabsecond}1.57M &\cellcolor{tabsecond}21.44 & \cellcolor{tabsecond}0.334 & \cellcolor{tabsecond}0.595 & \cellcolor{tabsecond}522.57 & \cellcolor{tabsecond}37.54 & \cellcolor{tabsecond}1.57M& \cellcolor{tabsecond}23.67 & \cellcolor{tabsecond}0.296 & \cellcolor{tabsecond}0.663 & \cellcolor{tabsecond}269.63& \cellcolor{tabsecond}34.99 \\

\textbf{SceneTok} & \cellcolor{tabfirst}\textbf{65.54K} & \cellcolor{tabfirst}\textbf{21.95} & \cellcolor{tabfirst}\textbf{0.219} & \cellcolor{tabfirst}\textbf{0.660} & \cellcolor{tabfirst}\textbf{246.51} & \cellcolor{tabfirst}\textbf{19.12} & \cellcolor{tabfirst}\textbf{32.77K} & \cellcolor{tabfirst}\textbf{24.21} & \cellcolor{tabfirst}\textbf{0.217} & \cellcolor{tabfirst}\textbf{0.683} & \cellcolor{tabfirst}\textbf{129.77} & \cellcolor{tabfirst}\textbf{13.94}\\
\bottomrule
    \end{NiceTabular}
    }
   \vspace{-0.3cm}

    \caption{\textbf{Quantitative comparison of NVS quality on DL3DV-140 and ACID}: \ours achieves better performance in most metrics as compared to baselines with explicit and latent representations on DL3DV-140 and superior generalization on zero-shot ACID. Note that the models used in ACID were trained on RealEstate10K (c.f. Tab.~\ref{tab:nvs_re10k}). 
    }
    \label{tab:nvs_dl3dv_acid}
    \vspace{-0.4cm}
\end{table*}
\vspace{-0.3cm}
\paragraph{Baselines.} We evaluate \ours, MVSplat~\cite{mvsplat}, MVSplat360~\cite{mvsplat360}, DepthSplat~\cite{depthsplat}, Long-LRM~\cite{longlrm}, RayZer~\cite{rayzer}, and LVSM~\cite{lvsm2025} on novel-view synthesis. Note that we mainly compare our method against LVSM, as it employs a latent representation. Although RayZer also employs a latent representation, it uses target views as inputs, unlike \ours and LVSM, which results in information leakage (c.f. Sec.~\ref{sec:generalizable_3d} and~\ref{sec:nvs}). For scene generation, we provide fair comparisons against RealEstate10K baselines, including DFM~\cite{dfm}, a 3D diffusion model with an underlying NeRF representation, and a video diffusion model DFoT~\cite{dfot}. We also compare against SEVA~\cite{seva}, which is trained on a large corpus of closed-source data.  
\begin{table}
    \centering
    \resizebox{\linewidth}{!}{
     \begin{NiceTabular}{l|ccc|ccc|ccc}
     \toprule
        \multicolumn{1}{c}{}
        & \multicolumn{3}{c}{R. Acc. $\uparrow$} 
        & \multicolumn{3}{c}{T. Acc. $\uparrow$}  
        & \multicolumn{3}{c}{AUC $\uparrow$} \\
         
        & $10^\circ$ & $20^\circ$ & $30^\circ$
        & $10^\circ$ & $20^\circ$ & $30^\circ$
        & $10^\circ$ & $20^\circ$ & $30^\circ$ \\ 
     \midrule
        
        \baseline{RayZer}~\cite{rayzer} & \baseline{\cellcolor{tabthird}26.16} & \baseline{\cellcolor{tabthird}42.50} & \baseline{\cellcolor{tabthird}54.45} & \baseline{\cellcolor{tabthird}07.82} & \baseline{\cellcolor{tabthird}23.75} & \baseline{\cellcolor{tabthird}41.59} & \baseline{\cellcolor{tabthird}0.007} & \baseline{\cellcolor{tabthird}0.020} & \baseline{\cellcolor{tabthird}0.096} \\
        
        LVSM~\cite{lvsm2025} &  \cellcolor{tabsecond}48.07 &  \cellcolor{tabsecond}65.60 &  \cellcolor{tabsecond}74.56 &  \cellcolor{tabsecond}14.25 &  \cellcolor{tabsecond}33.02 &  \cellcolor{tabsecond}49.60 &  \cellcolor{tabsecond}0.032 &  \cellcolor{tabsecond}0.103 &  \cellcolor{tabsecond}0.178 \\

        \textbf{SceneTok}  & \cellcolor{tabfirst}\textbf{75.81} & \cellcolor{tabfirst}\textbf{86.47} & \cellcolor{tabfirst}\textbf{89.86} & \cellcolor{tabfirst}\textbf{62.44} & \cellcolor{tabfirst}\textbf{80.31} & \cellcolor{tabfirst}\textbf{86.48} & \cellcolor{tabfirst}\textbf{0.307} & \cellcolor{tabfirst}\textbf{0.492} & \cellcolor{tabfirst}\textbf{0.593} \\
        
     \bottomrule
    \end{NiceTabular}
    }
    
    \vspace{-0.2cm}
   
    \caption{\textbf{Transferability on scene pairs from DL3DV-140.} We compare the transferability of novel camera trajectories by swapping target poses of scene pairs and quantify using TPS~\cite{mitchel2025true}. 
    }
    \label{tab:transferability}
    \vspace{-0.2cm}
\end{table}

\begin{table}
    \centering
    \resizebox{\linewidth}{!}{
     \begin{NiceTabular}{l|c|c|c}
     \toprule
        
        & $\text{MEt3R}_{RE10K} \downarrow$ & $\text{MEt3R}_{DL3DV} \downarrow$ & $\text{MEt3R}_{ACID} \downarrow$ \\
        \midrule

        \baseline{MVSplat360}~\cite{mvsplat360} & \baseline{\cellcolor{tabthird}0.0220} & \baseline{\cellcolor{tabthird}0.0579} & \baseline{\cellcolor{tabthird}0.0196} \\

        LVSM~\cite{lvsm2025} & \cellcolor{tabsecond}0.0162& \cellcolor{tabsecond}0.0541& \cellcolor{tabfirst}\textbf{0.0126}\\

        \textbf{SceneTok}  & \cellcolor{tabfirst}\textbf{0.0149}& \cellcolor{tabfirst}\textbf{0.0538}& \cellcolor{tabsecond}0.0133\\

     \bottomrule
    \end{NiceTabular}
    }
    
    \vspace{-0.2cm}
   
    \caption{\textbf{Multi-View Consistency.} \ours achieves good pairwise consistency with MEt3R~\cite{met3r} across all dataset.
    }
    \vspace{-0.5cm}
\end{table}
\vspace{-0.3cm}
\paragraph{Dataset.} We train and test on RealEstate10K~\cite{zhou2018stereo}, DL3DV~\cite{dl3dv}, and further experiment with zero-shot generalizability on ACID~\cite{acid}. For RealEstate10K, we evaluate in two settings where we select: (1) 270 scenes, each with 5 context and 64 target views, which we define as the narrow-view baseline; (2) 136 scenes, each with 12 context and 128 target views, which we define as the wide-view baseline. For DL3DV, we evaluate 140 scenes, each with 16 context and 64 target views; whereas on ACID, we follow the wide-view baseline as in (2). For scene generation, we select 200 scenes, each with 192 target views, where we use the first frame as the conditioning and select 12 anchor poses equally spaced across 192 frames (c.f. Sec.~\ref{sec:generator}). 

\vspace{-0.3cm}
\paragraph{Evaluation Metrics.} We evaluate the novel-view synthesis and generation quality using Peak-Signal-to-Noise Ratio (PSNR), Perceptual Similarity (LPIPS)~\cite{zhang2018perceptual}, Structural Similarity (SSIM), Frechet Video Distance (FVD)~\cite{unterthiner2019fvd}, and Frechet Inception Distance (FID)~\cite{fid}. Moreover, for multi-view consistency, we use MEt3R~\cite{met3r} and for transferability, we use the True-Pose-Similarity (TPS) metric proposed in concurrent work~\cite{mitchel2025true}.

\vspace{-0.3cm}
\paragraph{Training Details.} We train \ours on 4 Nvidia A100 (40GB) GPU with an effective batch size of 128 with gradient checkpointing, for 760K iterations ($\approx$ 11 epochs), resulting in a total of roughly $\approx250$ (GPU+CPU) hours. Furthermore, we train \ourscene for 1M iterations, which takes $\approx238$ (GPU+CPU) hours at an effective batch size of 96 on 4 Nvidia A100 (80 GB). Please refer to Appendix Sec.~\ref{sec:training_architecture_add} for more details.

\begin{figure*}[t]
    \centering

    \includegraphics[width=1\linewidth]{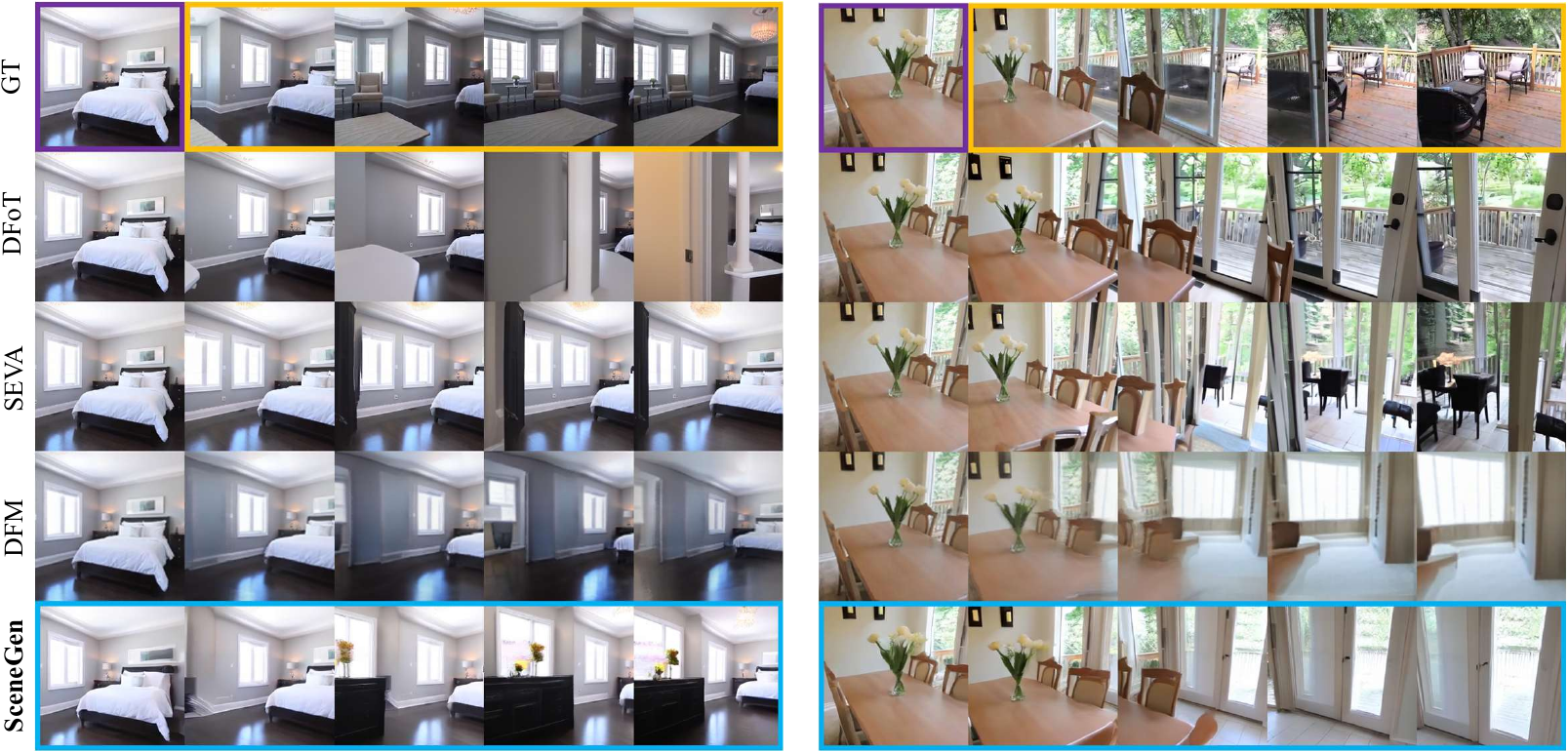}
    \vspace{-0.6cm}
    \caption{\textbf{Qualitative single-view generation comparison.} $($\textcolor{purple}{\thickBox}, \textcolor{ours}{\thickBox}, \textcolor{orange}{\thickBox}$)$ represents the input view, the rendered target views from the generated scene, and the ground-truth views respectively. DFM produces blurry results and struggles to generate complete structures. Although our method falls short in generating high-quality, high-frequency details when compared to DFoT, which performs pixel-space diffusion, and the foundational generative model SEVA, it is significantly more efficient (c.f. Tab.~\ref{tab:single_view_generation}). More examples in Appendix Sec.~\ref{sec:qualitative_results_add}. 
    }
    \label{fig:single_view_generation}
    \vspace{-0.4cm}
\end{figure*}

\subsection{Novel View Synthesis}
\label{sec:nvs}
In this section, we evaluate \ours on: (1) novel-view synthesis and (2) transferability of novel trajectories that deviate from the inputs. In (1), we compare existing works that we broadly categorize into \emph{explicit} and \emph{latent} representations. In (2), we only compare against the latter, as any classical 3D representation always performs true NVS. 

\vspace{-0.3cm}
\paragraph{Reconstruction Quality.}
In Tab.~\ref{tab:nvs_re10k} and~\ref{tab:nvs_dl3dv_acid}, we evaluate \ours and the baselines on RealEstate10K~\cite{zhou2018stereo}, DL3DV~\cite{dl3dv}, and zero-shot performance on ACID~\cite{acid}. We observe an overall improvement over the baselines in most metrics across all NVS settings while significantly reducing the representation size in terms of the number of 32-bit floating point numbers. Although Long-LRM~\cite{longlrm} achieves superior PSNR, LPIPS, and SSIM at the cost of a very large representation size, it falls short in rFVD and rFID. On the other hand, RayZer~\cite{rayzer} surpasses all baselines, including ours, in all metrics. However, Rayzer requires the target views as input, thus, cannot render novel trajectories (c.f.~\cite{mitchel2025true} and Tab.~\ref{tab:transferability}). Also, there might be basic statistics leaking through the pose encoder. DepthSplat achieves superior SSIM due to the use of extensive depth supervision during training and inference, whereas our method is self-supervised and is trained end-to-end with the rectified flow loss. Compared to our method, LVSM~\cite{lvsm2025} and RayZer use a dimensionality of $3072\times(512/768)$ respectively, which is significantly larger and ill-suited for generation with diffusion models. This is also true for all other 3D Gaussian-based baselines. We use a relatively light-weight video rectified flow model with 595M parameters, and we note that our decoder is efficient, attributed to the use of VideoDCAE~\cite{opensora2} and our compressed token representation. This allows us to render 32 novel views at a time in 1 second with 25 sampling steps. In Fig.~\ref{fig:qualitative_nvs}, we show qualitative comparisons of the rendered results on each dataset.

\begin{figure}[t]
    \centering
    \includegraphics[width=1.0\linewidth]
    {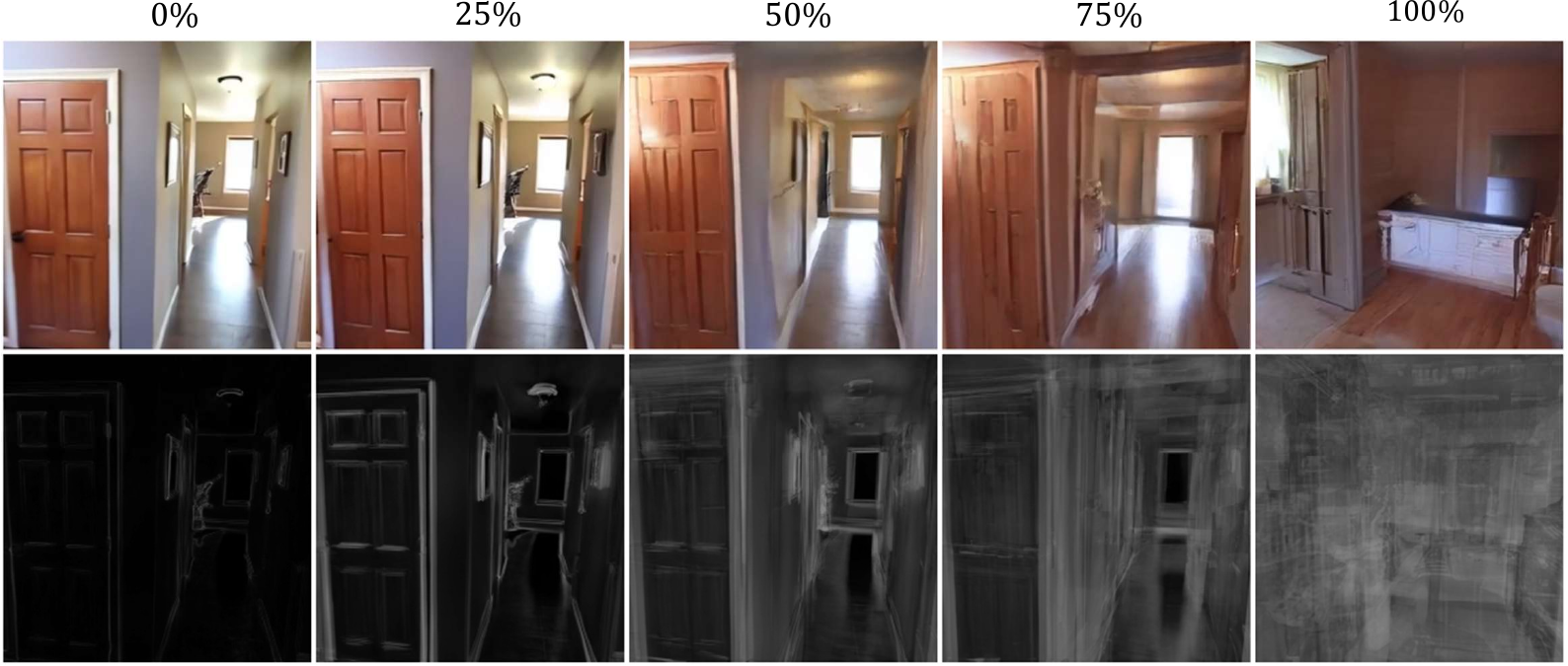} 
    \vspace{-0.5cm}
    \caption{\textbf{RGB and per-pixel variance vs mask-ratio.} We show a single RGB rendering with its variance map (over 10 samples) on increasing mask-ratios of $0\rightarrow100\%$. Note that we scale the variance map by a factor of 1.5.
    }
    \label{fig:variance_vs_mask}
    \vspace{-0.6cm}
\end{figure}
\vspace{-0.4cm}      
\paragraph{Transferability on Novel Trajectories.}
\label{par:transferability}
We further evaluate \ours on trajectories that deviate from the context trajectory in Tab.~\ref{tab:transferability}. We aim to verify that our method can generalize well to novel trajectories and perform true NVS~\cite{mitchel2025true} instead of view interpolation. We split DL3DV-140~\cite{dl3dv} scenes into 70 pairs, where we predict the scene tokens from the context views of scene A and render the target trajectory of scene B, while measuring pose accuracy with TPS~\cite{mitchel2025true} across 3 error thresholds. We find that our model follows the given trajectories more faithfully, as indicated by higher TPS, and surpasses LVSM~\cite{lvsm2025} and RayZer~\cite{rayzer}, which represent the state-of-the-art among generalizable 3D reconstruction methods without an explicit 3D representation. Please refer to Fig.~\ref{fig:transfer_1}, and~\ref{fig:transfer_2} in the appendix for qualitative comparisons. 

\begin{table}
    \centering
    \resizebox{\linewidth}{!}{
    \begin{NiceTabular}{l|ccc|cc|c|c}
     \toprule
    & PSNR $\uparrow$ & LPIPS $\downarrow$ & SSIM $\uparrow$ & gFVD $\downarrow$ & gFID $\downarrow$ &  Inference (s) $\downarrow$ & MEt3R $\downarrow$\\
     \midrule
        
        DFM~\cite{dfm} & \cellcolor{tabfirst}\textbf{15.86} & 0.468& \cellcolor{tabfirst}\textbf{0.529} &  566.71 & 52.64 & \cellcolor{tabthird}630 & 0.0225 \\
        DFoT~\cite{dfot} &  \cellcolor{tabthird}14.76 & \cellcolor{tabsecond}0.433 &  \cellcolor{tabthird}0.469 &  \cellcolor{tabthird}220.36 & \cellcolor{tabthird}35.40 & \cellcolor{tabsecond}146 & \cellcolor{tabthird}0.0194  \\
        SEVA~\cite{seva} &  14.41 & \cellcolor{tabthird}0.442 & 0.450 &  \cellcolor{tabfirst}\textbf{133.00} & \cellcolor{tabfirst}\textbf{17.69} & 1620 &\cellcolor{tabsecond}0.0164\\
        
     \midrule

        \textbf{SceneGen} &  \cellcolor{tabsecond}15.12 &  \cellcolor{tabfirst}\textbf{0.415} &  \cellcolor{tabsecond}0.493 &   \cellcolor{tabsecond}157.89 &  \cellcolor{tabsecond}18.90 & \cellcolor{tabfirst}\textbf{26 (11 + 16)} & \cellcolor{tabfirst}\textbf{0.0156}\\

     \bottomrule
        
    \end{NiceTabular}}
    \vspace{-0.2cm}
    \caption{\textbf{Single-View Generation Comparison.} We test \ourscene and the baselines on 200 scenes, each with 192 frames. \ourscene performs comparably to previous works in metrics but is significantly faster in generation ($\approx11s$).}
    \label{tab:single_view_generation}
    \vspace{-0.5cm}
\end{table}

\subsection{Scene Generation}
\label{sec:scene_generation}
We evaluate our latent scene generation model on the single-view generation task, and compare with DFM~\cite{dfm}, DFoT~\cite{dfot}, and SEVA~\cite{seva}, in terms of the FID and FVD, in addition to PSNR, LPIPS, and SSIM. Table~\ref{tab:single_view_generation} shows the quantitative results, and the qualitative results are provided in Fig.~\ref{fig:single_view_generation}. The quantitative results suggest that \ourscene is on par with large-scale multi-view generation models. Qualitatively, we can observe that DFoT and SEVA still generate slightly better high-frequency details. DFoT generates in pixel space without perceptual compression; therefore, it can produce high-frequency structures well. SEVA, on the other hand, has superior FID and FVD, which we attribute to the use of a large-scale closed-source dataset for training. In contrast, our method, DFM, and DFoT are all trained exclusively on RealEstate10K. Moreover, we achieve significant improvements in inference speed, as shown in Tab.~\ref{tab:single_view_generation}.  The inference speed is recorded on Nvidia H100 (80GB) for all methods and for \ourscene we show the combined inference time for generating the scene tokens ($11s$) and rendering 192 target views ($16$s). Unlike DFM, SEVA, and DFoT, which ran into OOM issues, \ourscene can also run on a local Nvidia RTX 4090 (24GB), with an even lower total inference time of $10s$ ($5s$ for generation and $5s$ for rendering 192 views).

\begin{figure}
    \centering
    
    \includegraphics[width=1\linewidth]{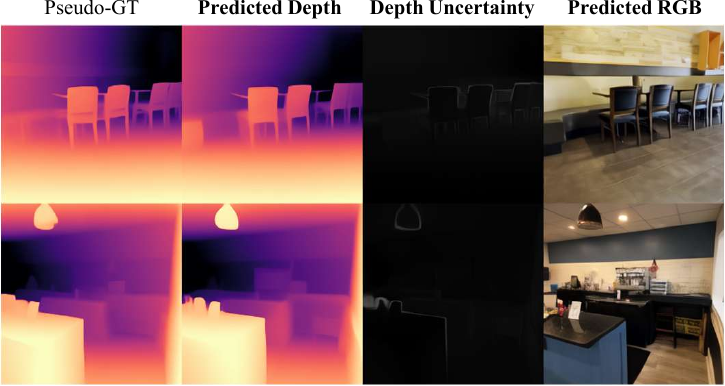}
    \vspace{-0.5cm}
    \caption{\textbf{Depth decoding on frozen SceneTok on DL3DV-140.} We show two scenes with the pseudo-GT from UniDepthV2~\cite{unidepthv2} and the predicted depth renderings with our finetuned depth decoder along with depth uncertainty and the RGB rendering. More examples are provided in Fig.~\ref{fig:depth_decoding_appendix} in the appendix.}
    \vspace{-0.4cm}
    \label{fig:depth_decoding}
\end{figure}

\subsection{Analysis on Scene Tokens}
\label{sec:analysis}
We perform further analysis on the token representation to understand what information is encoded and how it is provided to the decoder. We do this in three ways: (1) freezing the token space and decoding depth maps, (2) masking elements of the representation by dropping tokens, and (3) masking observations by dropping context views. In all experiments, we compute a per-pixel variance map of the output depth/RGB renderings across different noise initializations to visualize uncertainty. 
\vspace{-0.4cm}
\paragraph{Depth Decoding.} We use pre-trained \ours on DL3DV~\cite{dl3dv} and freeze the encoder. We replace the target images with depth maps predicted by UniDepthV2~\cite{unidepthv2} as pseudo ground-truth and compute video latents by duplicating the channels $\times 3$, and then fine-tune the pretrained decoder. In Fig~\ref{fig:depth_decoding}, we show that \ours inherently encodes geometric cues without any explicit supervision and can render plausible depth maps with minimal uncertainty. 

\vspace{-0.4cm}
\paragraph{Masking Scene Tokens.} We showcase the effect of randomly masking a percentage of the scene tokens, as shown in  Fig.~\ref{fig:variance_vs_mask}, which displays the RGB renderings and the corresponding per-pixel variance map for mask ratios of $0\%$, $25\%$, $50\%$, $75\%$, and $100\%$ (where we effectively perform unconditional sampling).  With 0\% masked tokens, the overall structure of the scene is well-encoded in the tokens, as indicated by low overall variance. In this case, the generative decoder only samples minor, high-frequency details around edges. By masking more tokens, we effectively remove information from the scene; therefore, we observe an increase in output variance.

\begin{figure}[t]
    \centering
    \includegraphics[width=1.0\linewidth]{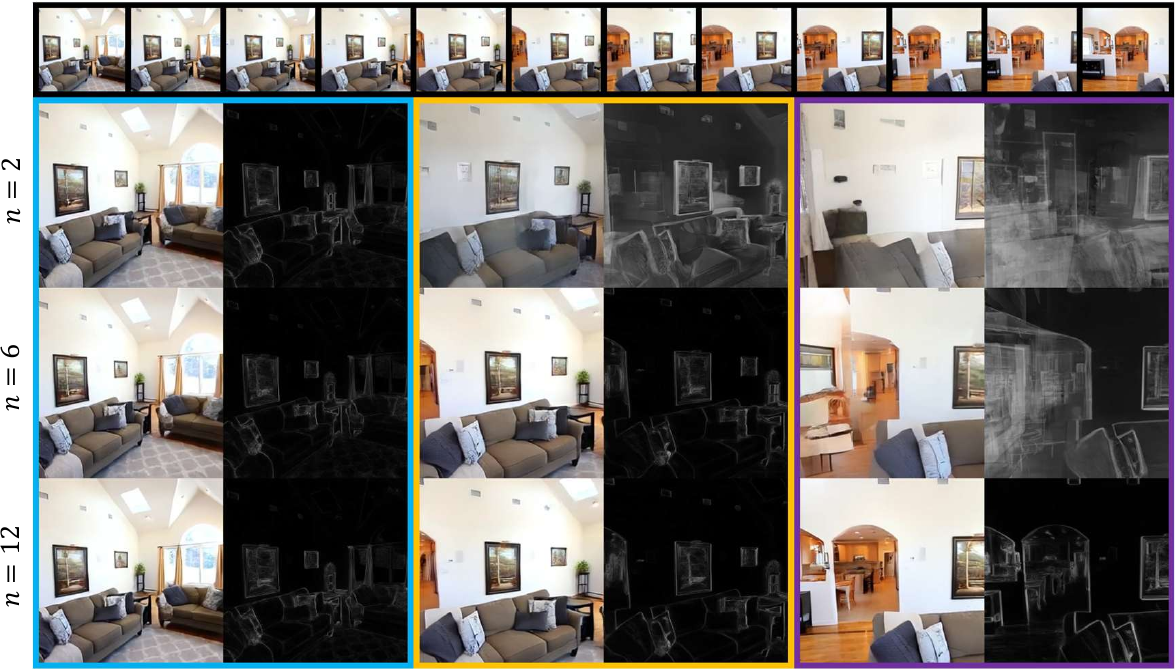} 
    \vspace{-0.5cm} 
    \caption{\textbf{RGB and per-pixel variance vs context views.} $($\textcolor{ours}{\thickBox}, \textcolor{orange}{\thickBox}, \textcolor{purple}{\thickBox}$)$ denote different target views and \thickBox~denotes the 12 context views. As we increase $n$, information in the tokens increases, and the variance reduces accordingly for the newly observed regions. Note that we scale the variance map by a factor of 2.}
    \label{fig:variance_vs_context}
    \vspace{-0.5cm}
\end{figure}

\vspace{-0.4cm}
\paragraph{Masking Context Views.} Alternatively, we drop the context views to visualize the effect more locally within the scene. Specifically, we keep the first $n$ out of 12 context views and drop the rest; this way, we progressively add/remove sections from the representation. In Fig.~\ref{fig:variance_vs_context}, we show the corresponding RGB rendering and the variance map for $n\in\{2, 6, 12\}$. We observe that, with increasing $n$, the output variance progressively reduces, and regions covered by the tokens exhibit lower variance. This shows that our scene token plays a crucial role in providing most of the information to the decoder. We provide further quantitative analysis in Appendix Sec.~\ref{sec:scene_token_analysis_add}.

\section{Conclusion}
We introduced \ours, an autoencoder to encode scene information from view sets into an unstructured, highly compressed token space. The architecture consists of a scene perceiver module and a generative decoder. We demonstrated that the autoencoder reaches state-of-the-art reconstruction quality, is very efficient, and decodes from an order-of-magnitude stronger compression. We demonstrated that the latent space lends itself to efficient scene generation in 5 seconds on a single consumer GPU.

\vspace{-0.4cm}
\paragraph{Limitations and Future Work.} \ours introduces a novel concept for scene encoding but still has limitations, for example in reconstructing consistent high-frequency details. We see potential for improvements with better image compressors or through structured latent spaces, utilizing findings from recent works on image tokenization~\cite{flextok, yu2025repa}. 

\section*{Acknowledgements}
This project was partially funded by the Saarland/Intel Joint Program on the Future of Graphics and Media.
Jan Eric Lenssen is supported by the German Research Foundation (DFG) - 556415750 (Emmy Noether Programme, project: Spatial Modeling and Reasoning).

{
    \small
    \bibliographystyle{ieeenat_fullname}
    \bibliography{main}
}
\clearpage
\setcounter{page}{1}
\begin{appendices}
\maketitlesupplementary

The supplementary materials are structured as follows.
First, we provide detailed information about our training procedure and the architecture for both \ours and \ourscene in Sec.~\ref{sec:training_architecture_add}, ablation on the design choices for both models in Sec.~\ref{sec:ablation}, a discussion on the choice of positional encoding, visualization of the latent space and further analysis into the uncertainty caused by adding or removing information from the scene tokens in Sec.~\ref{sec:scene_token_analysis_add}, and additional qualitative results for transferability on novel trajectories, novel-view synthesis, and single-view generation in Sec.~\ref{sec:qualitative_results_add}.  

\section{Training and Architectural Details}
\label{sec:training_architecture_add}
\subsection{SceneTok.} 
\begin{figure}
    \centering
    \includegraphics[width=0.8\linewidth]{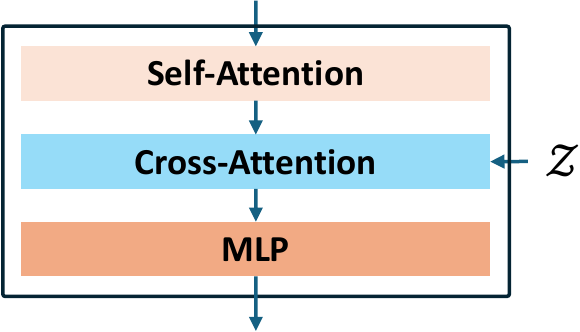}
    \caption{\textbf{Decoder block of \ours.}}
    \label{fig:decoder_block}
    \vspace{-0.3cm}
\end{figure}

In Tab.~\ref{tab:scenetok_hyp}, details of the model and training hyper-parameters, including the input and sampling configuration, are provided. In addition, Fig.~\ref{fig:decoder_block} shows a single decoder block, where we first perform self-attention between latent (noisy) video tokens of the target views, followed by cross-attention to the scene tokens $\mathcal{Z}$.  

\vspace{-0.4cm}
\paragraph{Choice of Input Coordinate System.}
When encoding 3D scenes from view sets, we choose one input view as reference and ensure that all other view poses are given as relative pose w.r.t. to the chosen reference pose. The choice of reference frame can vary and can be augmented during training.  The procedure ensures that the scene origin is always present in the input and that the scene is structured around this origin, improving model generalization.
Anchoring the representation around the origin is also relevant for scene generation, as further discussed in Sec.~\ref{sec:generator}.

\vspace{-0.4cm}
\paragraph{Camera-Controlled Video Rendering.} We use VideoDCAE~\cite{opensora2} (later also add WanVAE~\cite{wan2025} c.f. Sec.~\ref{sec:ablation}) as the tokenizer for the target views which allows for 32x spatial and 4x temporal compression. We assume each latent video frame is temporally localized and encodes four consecutive RGB frames. Therefore, we concatenate the corresponding ray maps along the channel dimension for these four frames. The high spatio-temporal compression allows us to render more frames quickly and efficiently.

\vspace{-0.4cm}
\paragraph{Data Augmentation.} During training, we randomly choose our reference frame from the $N$ context views and compute the corresponding relative camera poses for the remaining context and target views. We also randomly flip images and their trajectories, as done in PixelSplat~\cite{pixelsplat}. 

\vspace{-0.4cm}
\paragraph{Training Instability.} We observe some training instability (gradient spikes) during our training. We attribute this primarily to the use of mixed-precision, and it mainly occurs during the encoding stage. To assist with the training, we use a relatively lower gradient clipping value, skip the optimizer step when large gradient norms are detected, following ~\cite{lvsm2025, rayzer}, later freeze the encoder, and continue finetuning the decoder on the frozen latent space.   

\begin{table}
    \centering
    \resizebox{\linewidth}{!}{
    \begin{tabular}{l|c|c}
    \toprule

        \emph{Model Configurations} & Encoder & 
        Decoder\\
       \midrule 
        Overall size & $242M$ & $595M$ \\
        Hidden dimensions $d_H$ & $768$ & $1024$ \\
        Depth $L$ & $12$ & $24$\\
        Number of heads & $12$ & $16$ \\
        MLP ratio & $4.0$ & $4.0$ \\
        2D RoPE & \cmark & \xmark \\
        3D RoPE & \xmark & \cmark \\
        Norm type & RMSNorm & RMSNorm \\
        Camera conditioning & AdaLN & AdaLN \\
        QK Norm & \xmark & \cmark \\
        Activation & SwiGLU & SwiGLU \\
        \midrule
        \emph{Inputs Configurations}\\
       \midrule 
        Raw shape & $N\times256\times256\times3$ & $M\times256\times256\times3$ \\
        Temporal compression & $\times1$ & $\times4$ \\
        Spatial compression & $\times16$ & $\times32$ \\
        Token dimensions & $32$ & $128$ \\
        Patch size & $2$ & $1$\\
        \midrule
       \emph{Training Configurations} & RealEstate10K & DL3DV\\
       \midrule 
        Optimizer & AdamW & AdamW \\
        Initial LR &  $1.0e^{-4}$ & $1.0e^{-4}$ \\
        LR decay steps &  $[20K, 150K, 500K]$ & $[4K, 8K, 100K]$ \\
        LR decay factor & $0.4$ & $0.4$ \\
        Warmpup steps & $2000$ & $2000$ \\
        Gradient clip & $0.01$ & $0.01$ \\
        Precision & bfloat16 (mixed) & bfloat16 (mixed) \\
        Effective Batch size & $128$ & $64$ \\
        Training steps & $760K$ & $1M$ \\
        Freeze encoder after & $150K$ & None \\
        \midrule
       \emph{Token Dimensionality}\\
       \midrule 
        RealEstate10K / ACID &  \multicolumn{2}{c}{$512\times64$} \\
        DL3DV &  \multicolumn{2}{c}{$1024\times64$} \\
        \midrule 

        \emph{Sampling (Inference)}  \\
       \midrule 
        Steps & \multicolumn{2}{c}{$25$} \\
        CFG scale & \multicolumn{2}{c}{$1.0$} \\
        \bottomrule

    \end{tabular}
    }
    \caption{\textbf{Summary of \ours.} We show the model, input, training and sampling our scene autoencoder. Both the encoder and the decoder using the transformer modules from LightningDiT~\cite{yao2024fasterdit} with additional modifications (c.f. Fig.~\ref{fig:encoder_block} and Sec.~\ref{sec:method}).}
    \label{tab:scenetok_hyp}
\end{table}

\subsection{SceneGen}
During training, we use the encoder from \ours to predict the scene tokens $\mathcal{Z}$ from the given context views $\mathcal{X}_C$. We then add noise to the scene tokens and denoise them with \ourscene. Fig.~\ref{fig:decoder_block} shows the decoder block of \ours, which we reuse in \ourscene by replacing the keys and values, i.e., $\mathcal{Z}$ with $[\mathcal{F}_I, \mathcal{A}]$, which is a concatenation of the input image tokens (from VA-VAE~\cite{vavae}) and the anchor tokens along the sequence dimension. The scene tokens serve as the inputs and outputs of our latent generation model.

\begin{figure}
    \centering
    \includegraphics[width=1.0\linewidth]{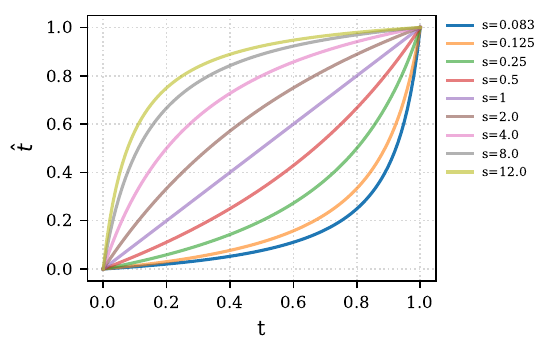}
    \vspace{-0.7cm}
    \caption{\textbf{Noise-level shift.} We plot the resulting timestep $\hat{t}$ for different amount of $s$. For $s>1$, we shift more towards higher noise and vice versa. }
    \label{fig:shift}
    \vspace{-0.4cm}
\end{figure}

In Tab.~\ref{tab:scenegen_hyp}, details of the model and training hyper-parameters, including the input and sampling configuration, are provided. Note that we use a higher timestep shift, which we found to improve the generation results. Specifically, we shift our timestep $t \rightarrow \hat{t}$ as follows:
\begin{equation}
    \hat{t} = \frac{st}{1+(s-1)t} \textnormal{,}
\end{equation}
where $s$ is the shift amount ($\uparrow$ higher and $\downarrow$ lower noise-levels). Refer to Sec.~\ref{sec:ablation} for further quantitative ablation over different shifts. In Fig~\ref{fig:shift}, we show the resulting shifted timestep $\hat{t}$ for $s>1$ and $s<1$.

\begin{table}
    \centering
    \resizebox{\linewidth}{!}{
    \begin{tabular}{l|c}
    \toprule

        \emph{Model Configurations}\\
       \midrule 
        Overall size & $1.3B$ \\
        Hidden dimensions $d_H$  & $1536$ \\
        Depth $L$ & $24$\\
        Number of heads  & $24$ \\
        MLP ratio  & $4.0$ \\
        Norm type  & RMSNorm \\
        Camera conditioning  & AdaLN \\
        Anchor conditioning & CrossAtten \\
        QK Norm  & \cmark \\
        Activation & SwiGLU \\
        \midrule
        \emph{Inputs Configurations}\\
       \midrule 
        Raw shape & $512\times64$ \\
        Token dimensions &64 \\
        \midrule
       \emph{Training Configurations}\\
       \midrule 
        Optimizer & AdamW \\
        Initial LR & $1.0e^{-4}$ \\
        LR decay steps & [100K, 250K, 400K] \\
        LR decay factor &0.4 \\
        Warmpup steps &2000 \\
        Gradient clip &0.1 \\
        Precision & bfloat16 (mixed) \\
        Effective Batch size & 96 \\
        Timestep weighting & Shifted Timestep \\
        Timestep shift & 12 \\
        Training steps & 1M \\
        \midrule
                
        \emph{Sampling (Inference)} \\
       \midrule 
        Steps & 150 \\
        CFG scale & 3.0 \\
        Timestep shift & 12 \\
        \bottomrule

    \end{tabular}
    }
    \caption{\textbf{Summary of \ourscene.} We show the model, input, training and sampling configurations for both versions of latent scene generation model. We use the transformer module from LightningDiT~\cite{yao2024fasterdit} with an additional cross attention module for input conditioning (c.f. Sec.~\ref{sec:generator}).}
    \vspace{-0.4cm}
    \label{tab:scenegen_hyp}
\end{table}

\vspace{-0.3cm}

\paragraph{Data Augmentation.} During training, we augment our scene origin by randomly selecting a single context view as the reference. The corresponding anchors and conditioning camera are set relative to the reference frame (before predicting the scene tokens). The reference view is also used as the first input conditioning for the model. We also randomly flip the images and poses horizontally, similar to \ours.  
\begin{figure}
    \centering
    \includegraphics[width=1.0\linewidth]{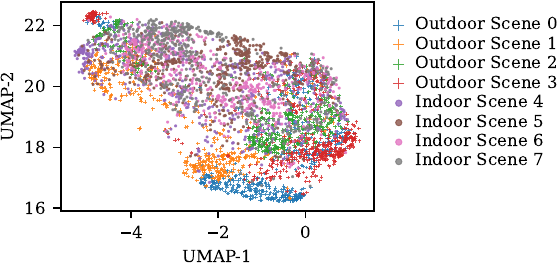}
    \vspace{-0.7cm}
    \caption{\textbf{UMAP plot of scene tokens across 8 scenes from RealEstate10K. } We compute the UMAP across all 512 tokens for 8 different scenes out of which 4 are outdoor ($+$) and indoor ($\bullet$) scenes (c.f. Fig~\ref{fig:indoor_outdoor_add}). We see that the majority of the tokens that belongs to the outdoor scenes are grouped closely together as compared to tokens from indoor scenes. This shows that our latent token space is discriminative in nature which is helpful for downstream generation task~\cite{vavae}.}
    \label{fig:umap_add}
    \vspace{-0.1cm}
\end{figure}
\begin{figure}
    \centering
    \includegraphics[width=1.0\linewidth]{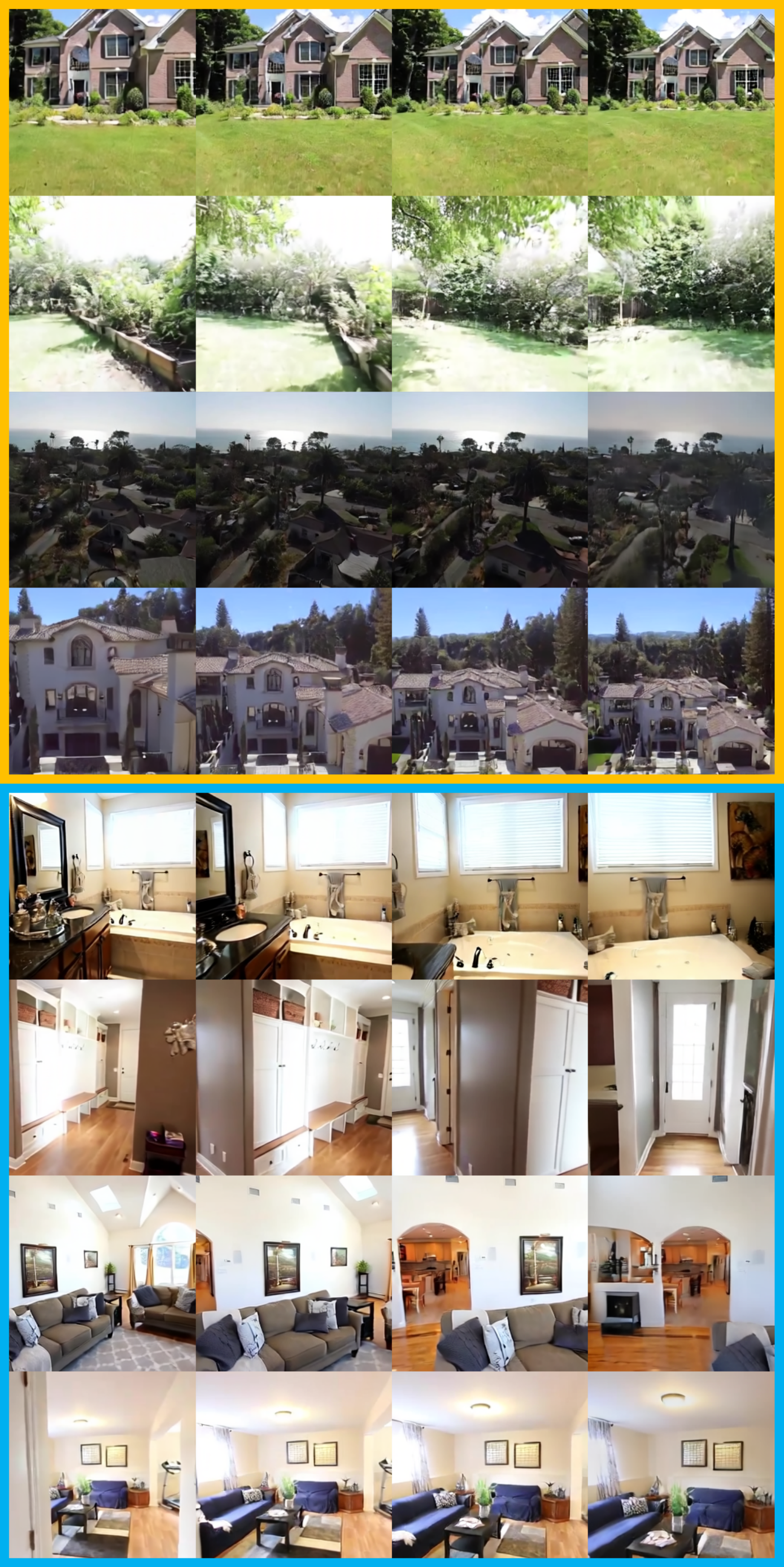}
    \vspace{-0.6cm}
    \caption{\textbf{Examples of outdoor and indoor scenes. }  (\textcolor{orange}{\thickBox}, \textcolor{ours}{\thickBox}) denote outdoor and indoor scenes respectively. We show the RGB renderings of the scenes used in Fig.~\ref{fig:umap_add}. Note that the scenes are ordered from top-to-bottom with indices $0\rightarrow7$. }
    \label{fig:indoor_outdoor_add}
    \vspace{-0.6cm}
\end{figure}

\begin{figure*}[t]
    \centering

    \includegraphics[width=1.0\linewidth]{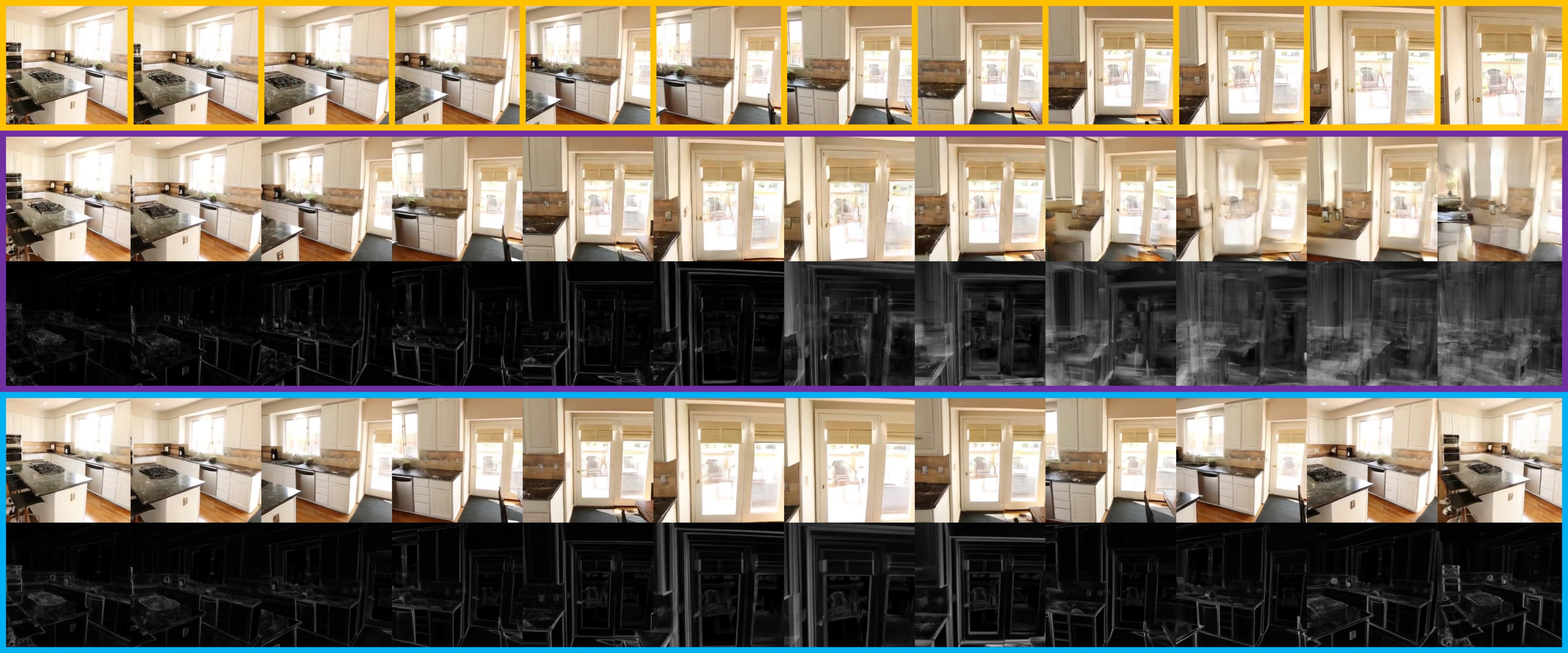}
    \vspace{-0.4cm}
    \caption{\textbf{Comparison of 3D vs 2D RoPE in the encoder.} \textcolor{orange}{\thickBox} denotes the input context views, $($\textcolor{purple}{\thickBox},  \textcolor{ours}{\thickBox}$)$ denotes the rendered targets with 3D RoPE~\cite{rope3d} and 2D RoPE~\cite{rope} respectively. We show 12 target views (RGB and variance map) from left-to-right where the first 6 columns shows the rendered views for the forward tajectory which aligns with the input order of the context and the remaining 6 views shows the backward trajectory. We observe higher variance for the backward trajectory for 3D RoPE and this indicates temporal bias of the scene tokens $\mathcal{Z}$. Note that we scale our variance map by a factor of 2.
    }
    \label{fig:cycle_add}
    \vspace{-0.3cm}
\end{figure*}

\section{Ablation Studies}
\label{sec:ablation}
In this section, we provide further results where we ablate specific design choices of our method, including both \ours and \ourscene. 

\vspace{-0.3cm}
\paragraph{Pretrained Video VAEs in \ours} We ablate the choice of the pre-trained image latent space (VideoDCAE~\cite{opensora2} in the main paper) by additionally evaluating WanVAE~\cite{wan2025} as replacement for VideoDCAE. We found that WanVAE achieves better high frequency details, which we attribute to its relatively lower spatial compression, i.e., $\times16$, as compared to $\times32$ in VideoDCAE and is indicated by superior rFID and rFVD in Tab.~\ref{tab:vae_ablation}. However, it achieves slightly worse results in reconstruction metrics, which are sensitive to blur and camera inaccuracies. Refer to the qualitative examples in Sec.~\ref{sec:qualitative_results_add} where we provide results for both versions of \ours.
\begin{table}
    \centering
    \resizebox{\linewidth}{!}{
     \begin{NiceTabular}{c|p{0.3\linewidth}|ccc|cc} 
     \toprule
    & \centering Compression \centering(Spatial, Temp.) & PSNR $\uparrow$ & LPIPS $\downarrow$ & SSIM $\uparrow$ & rFVD $\downarrow$ & rFID $\downarrow$  \\ \midrule

   VideoDCAE~\cite{opensora2} & $\times32$,$\times4$ & \cellcolor{tabfirst}\textbf{21.95} & \cellcolor{tabfirst}\textbf{0.219} & \cellcolor{tabfirst}\textbf{0.660} & \cellcolor{tabsecond}246.51 & \cellcolor{tabsecond}19.12 \\
    WanVAE~\cite{wan2025}) & $\times16$,($\times4 + 1)$  & \cellcolor{tabsecond}19.81 & \cellcolor{tabsecond}0.254 & \cellcolor{tabsecond}0.561 & \cellcolor{tabfirst}\textbf{189.74} & \cellcolor{tabfirst}\textbf{14.30} \\

     \bottomrule
    \end{NiceTabular}
    }
   
    \caption{\textbf{Ablation on pretrained Video VAEs in \ours decoder.} We achieve better high frequency details with the \ours renderer trained with WanVAE~\cite{wan2025}. This partly due to smaller spatial compression as compared to VideoDCAE. From this, we conclude that \ours performance is bounded by the performance of the video VAEs and therefore, exploring better alternatives can potentially improve \ours. 
    }
    \label{tab:vae_ablation}
    \vspace{-0.4cm}
\end{table}
\vspace{-0.3cm}

\paragraph{Classifier-Free Guidance in \ours} We notice that even though we train our decoder both conditionally and unconditionally to allow classifier-free guidance~\cite{ho2021classifierfree} during NVS, we found that using a higher guidance scale reduces the quality, as shown in Tab.~\ref{tab:cfg_ablation}, causing over-saturated RGB renderings and, therefore, worse reconstruction metrics. From these insights, we later remove classifier-free guidance when training \ours on DL3DV and instead train it purely conditionally.  
\begin{table}
    \centering
    \resizebox{\linewidth}{!}{
     \begin{NiceTabular}{c|ccc|cc} 
     \toprule
   Guidance Scale & PSNR $\uparrow$ & LPIPS $\downarrow$ & SSIM $\uparrow$ & rFVD $\downarrow$ & rFID $\downarrow$  \\ \midrule

   1.0 & \cellcolor{tabsecond}24.22 &\cellcolor{tabsecond} 0.157 & \cellcolor{tabthird}0.784 &  \cellcolor{tabfirst}68.39 &  \cellcolor{tabfirst}9.13 \\
   2.0 & \cellcolor{tabfirst}24.35 & \cellcolor{tabfirst}0.151 & \cellcolor{tabfirst}0.792 &  \cellcolor{tabsecond}69.52 & \cellcolor{tabsecond}9.69 \\
   3.0 & \cellcolor{tabthird}24.05 & \cellcolor{tabthird}0.158 & \cellcolor{tabsecond}0.787 &  \cellcolor{tabthird}79.60 &  \cellcolor{tabthird}11.25 \\
   4.0 & 23.71 & 0.166 & 0.778 &  89.11 &  12.75 \\
   5.0 & 23.49 & 0.174 & 0.771 &  101.24 &  14.19 \\
   6.0 & 23.28 & 0.183 & 0.764 &  112.04 &  16.00 \\
   7.0 & 23.06 & 0.190 & 0.757 &  124.79 &  17.375 \\

     \bottomrule
    \end{NiceTabular}
    }
   
    \caption{\textbf{Ablation on CFG scales in \ours.} We show the quantitative results for increasing guidance scale when rendering novel views. The experiment is performed on the wide-view baseline (c.f. Sec.~\ref{sec:experimental_setup}) using 12 context views and 128 target views. We clearly see that using a smaller guidance scale performs better. 
    }
    \label{tab:cfg_ablation}
\end{table}

\begin{figure}
    \centering
    \includegraphics[width=0.8\linewidth]{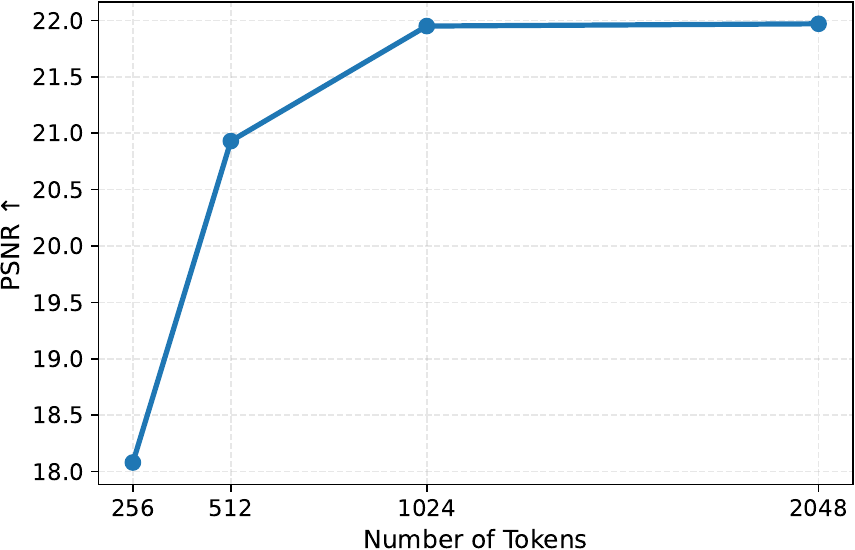}
    \vspace{-0.2cm}
    \caption{\textbf{PSNR vs Number of Tokens on DL3DV. } We train four variant of our method on different number of tokens while keeping the channel dimensions fixed.}
    \label{fig:psnr_vs_tokens}
    \vspace{-0.4cm}
\end{figure}

\vspace{-0.3cm}
\paragraph{Number of Tokens vs PSNR in \ours. } In Fig.~\ref{fig:psnr_vs_tokens}, we show the performance of our model trained on an increasing number of scene tokens and observe that with more tokens, we achieve better reconstruction quality; however, we chose $K=1024$, and increasing it further leads to early overfitting, and the quality saturates. We believe that our method becomes upper bounded by the reconstruction quality of the pretrained video VAEs.    

\vspace{-0.3cm}
\paragraph{Timestep Shift in \ourscene} We train the generative model with varying timestep shifts $s\in{1.0, 4.0, 12.0}$. Overall, we found that shifting the noise-levels towards higher noise improves the results in terms of gFID in Fig.~\ref{tab:shift_ablation}, suggesting that global structures emerge early on; therefore, it requires more steps to ensure better generation quality.

\section{Analysis on Scene Tokens}
\label{sec:scene_token_analysis_add}
\paragraph{Visualization of scene tokens.} In Fig.~\ref{fig:umap_add}, we show the UMAP~\cite{umap} plot of the scene tokens across 4 outdoor and 4 indoor scenes (refer to Fig~\ref{fig:indoor_outdoor_add}). The goal is to understand the structure of the latent space in terms of its discriminativeness; i.e., tokens that belong to similar scenes (outdoor) should be clustered closely together, and vice versa. From the figure, we observe that the majority of the tokens belonging to outdoor scenes are grouped more compactly, and the inter group distance (especially between outdoor and indoor scene tokens) is large; i.e., we observe a clear separation between them.  
\begin{table}
    \centering
    \resizebox{0.5\linewidth}{!}{
     \begin{NiceTabular}{c|c} 
         \toprule
       Timestep Shift (s) & gFID $\downarrow$  \\ \midrule
    
       1.0 & \cellcolor{tabthird}19.99  \\
       4.0 & \cellcolor{tabsecond}19.04 \\
       12.0 & \cellcolor{tabfirst}18.90 \\
    
         \bottomrule
        \end{NiceTabular}
    }
   
    \caption{\textbf{Ablation on Timestep Shift in \ourscene.} We show the gFID three different timestep shifts where we train a separate model for each shift amount and also apply it during inference. We found that shifting towards higher noise results in better global structures as more steps are allocated in the low frequency regime. 
    }
    \label{tab:shift_ablation}
    \vspace{-0.4cm}
\end{table}

\begin{figure}
    \centering
    \includegraphics[width=1.0\linewidth]{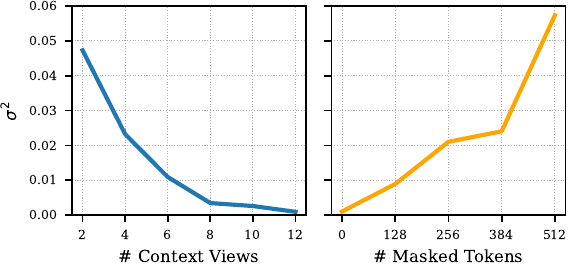}
    \caption{\textbf{Average per-pixel variance.} We plot the per-pixel variance averaged across the spatial and temporal dimension against increasing number of context views and masked tokens (out of 512 tokens). We show the plot for a single example scene.}
    \label{fig:variance_vs_mask_context_add}
    \vspace{-0.4cm}
\end{figure}
\vspace{-0.4cm}
\paragraph{Temporal biases in the scene tokens.} We experimented with 2 choices of positional encoding for the input context view tokens: 1) 3D RoPE~\cite{rope3d} and 2) 2D RoPE~\cite{rope}. In the former, we assume a temporally ordered sequence of context view tokens $\mathcal{X}_C=(\mathbf{x}_i\in\mathbb{R}^d)_{i=1}^N$ for each spatial location $(i, j)$. We perform a cycle consistency test, during which we render the target trajectory in both forward and backward order. In Fig.~\ref{fig:cycle_add}, we show RGB renderings and the corresponding per-pixel variance maps for the forward and backward trajectory, where we can see that with 3D RoPE, we observe higher variance and poor renderings in the backward trajectory. The corresponding RGB renderings for the backward trajectory fail to follow the given camera poses correctly. In contrast, with 2D RoPE, we achieve significantly better cycle consistency, indicating invariance to the input order.

\vspace{-0.4cm}
\paragraph{Quantitative analysis of uncertainty.} In Fig.~\ref{fig:variance_vs_mask_context_add}, we compute the per-pixel variance across 10 samples and average across the spatial and temporal dimensions. We observe a clear decreasing trend when we add more information to the scene tokens by progressively including more contextual views. Likewise, we observe a clear increasing trend when we remove information by masking the scene tokens.

\section{Additional Qualitative Results}
\label{sec:qualitative_results_add}  

\begin{figure*}[t]
    \centering
    \begin{subfigure}[t]{0.49\linewidth}
        
        \includegraphics[width=1.0\linewidth]{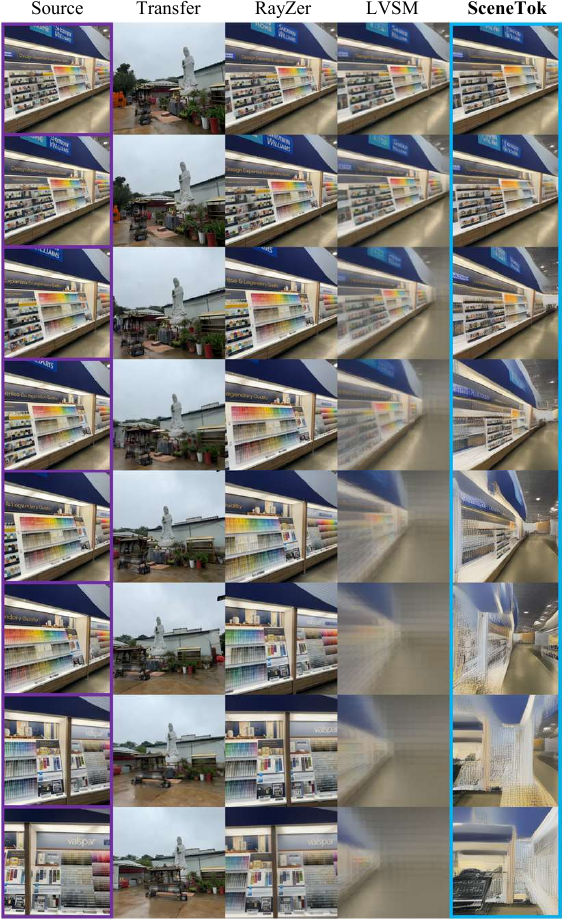}
    \end{subfigure}
    \begin{subfigure}[t]{0.49\linewidth}
        
        \includegraphics[width=1.0\linewidth]{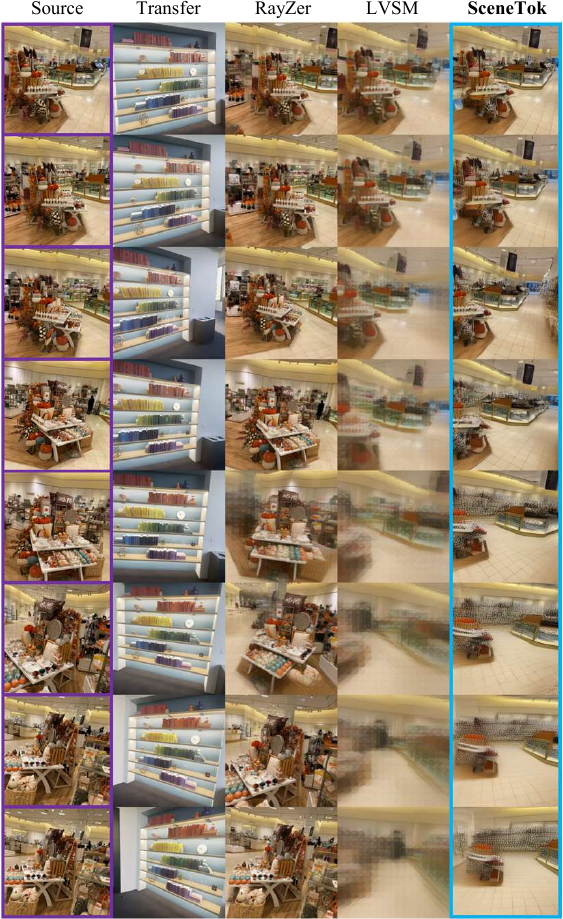}
    \end{subfigure}

    \caption{\textbf{Qualitative NVS on Transfer Trajectories.} $($\textcolor{purple}{\thickBox}, \textcolor{ours}{\thickBox}) denotes the context views of the \emph{source} scene, and the renderings of the \emph{source} scene with trajectory from the \emph{transfer} scene with \ours. Note that we can expect artifacts in regions that are not present in the context views and both \ours, and LVSM~\cite{lvsm2025} are not trained for extrapolation. RayZer on the other hand fails to follow the given trajectory and only renders the original trajectory of the \emph{source} scene. 
    }
    \label{fig:transfer_1}
    \vspace{-0.4cm}
\end{figure*}

\begin{figure*}[t]
    \centering
    \begin{subfigure}[t]{0.49\linewidth}
        
        \includegraphics[width=1.0\linewidth]{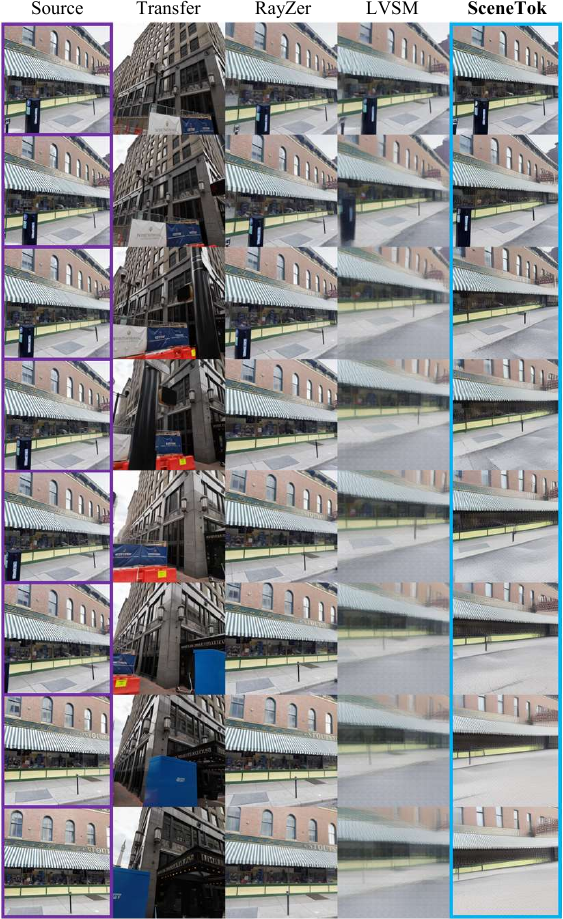}
        
    \end{subfigure}
    \begin{subfigure}[t]{0.49\linewidth}
        
        \includegraphics[width=1.0\linewidth]{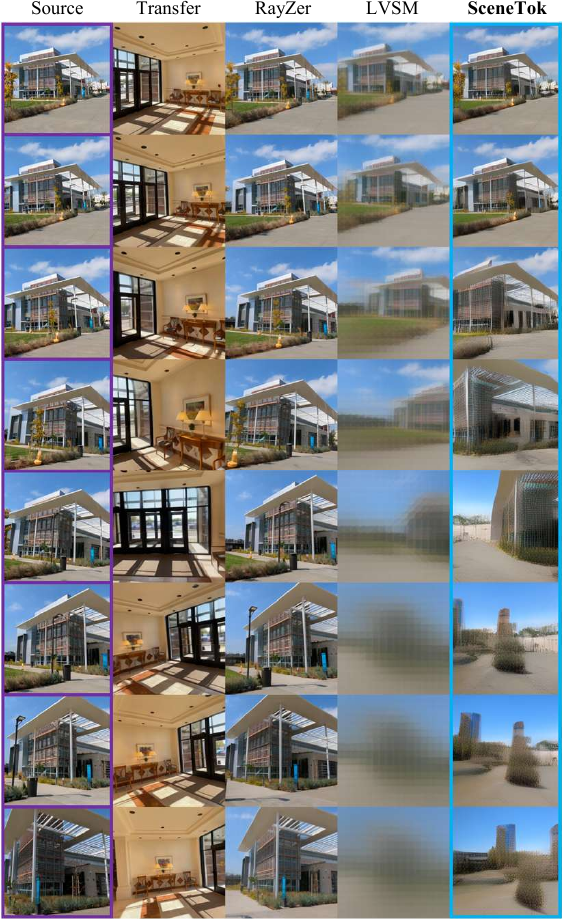}
    \end{subfigure}

    \caption{\textbf{Qualitative NVS on Transfer Trajectories.} $($\textcolor{purple}{\thickBox}, \textcolor{ours}{\thickBox}) denotes the context views of the \emph{source} scene, and the renderings of the \emph{source} scene with trajectory from the \emph{transfer} scene with \ours. Note that we can expect artifacts in regions that are not present in the context views and both \ours, and LVSM~\cite{lvsm2025} are not trained for extrapolation. RayZer on the other hand fails to follow the given trajectory and only renders the original trajectory of the \emph{source} scene. 
    }
    \label{fig:transfer_2}
    \vspace{-0.4cm}
\end{figure*}

\paragraph{Transferability on scene pairs.} In Fig.~\ref{fig:transfer_1} and~\ref{fig:transfer_2}, we provide qualitative visualizations of the transferability of novel trajectories that belong to a different scene. Specifically, consider the scene $A$ (source) and $B$ (transfer), we predict the scene tokens $\mathcal{Z}$ from the context views of scene $A$, and render the trajectory from scene $B$. We show that, with our method, we can perform novel view rendering. In our experiments, we ensure that the center camera position and orientation of both trajectories from scene $A$ and $B$ are aligned at the same origin with respect to the first camera pose. 

\begin{figure}
    \centering
    \includegraphics[width=1.0\linewidth]{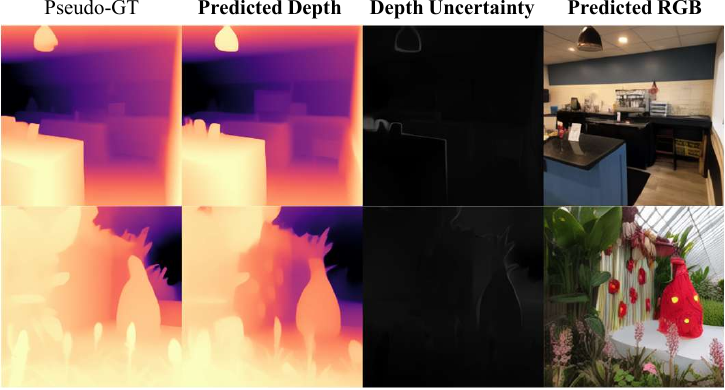}
    \caption{\textbf{Depth decoding on frozen SceneTok on DL3DV-140.} We show two scenes with the pseudo-GT from UniDepthV2~\cite{unidepthv2} and the predicted depth renderings with our finetuned depth decoder along with depth uncertainty and the RGB rendering.}
    \label{fig:depth_decoding_appendix}
    \vspace{-0.4cm}
\end{figure}

\vspace{-0.3cm}

\begin{figure*}[t]
    \centering

    \includegraphics[width=1.0\linewidth]{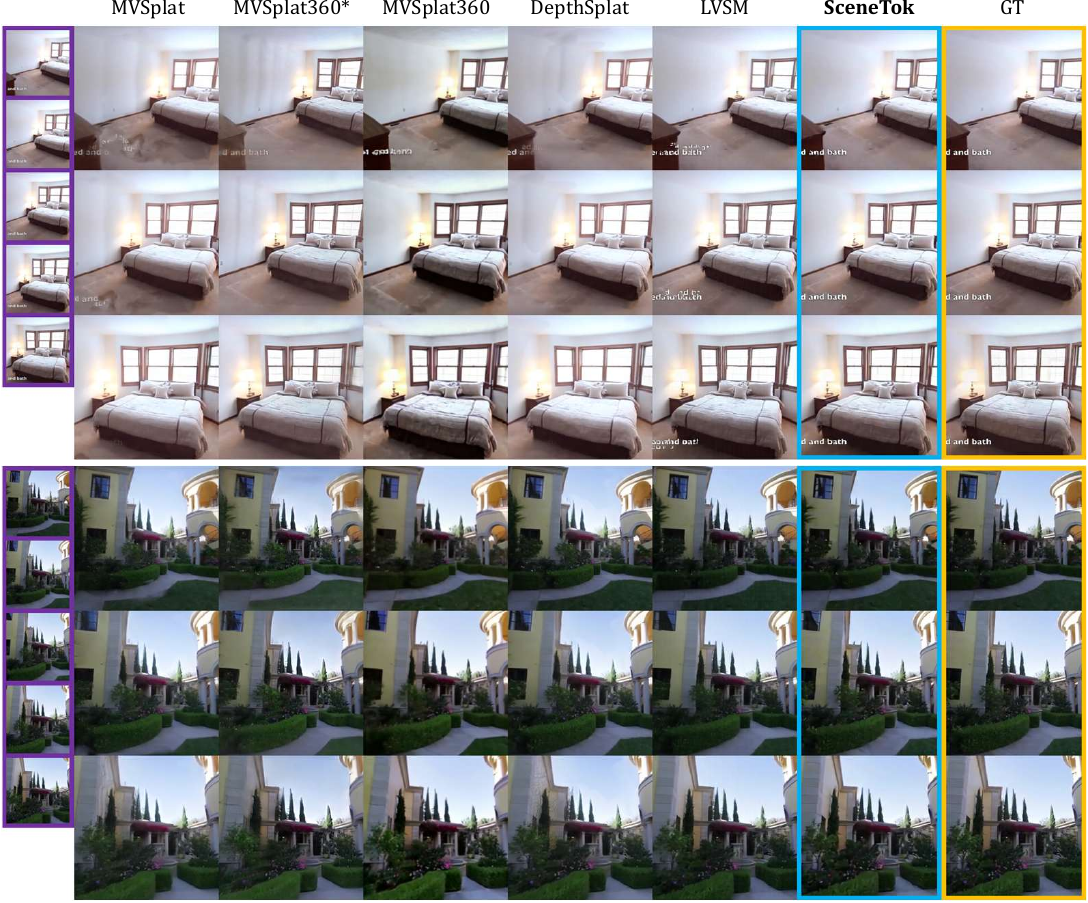}
    \vspace{-0.4cm}
    \caption{\textbf{Qualitative NVS on RealEstate10K.} $($\textcolor{purple}{\thickBox}, \textcolor{ours}{\thickBox}, \textcolor{orange}{\thickBox}$)$ denotes the 5 input context views, the target renderings from \ours and the corresponding ground-truth target views respectively (we show six views from top-to-bottom). * denotes output of MVSplat360 taken before the refinement step with a video diffusion model. 
    }
    \label{fig:qualitative_nvs_1}
    \vspace{-0.4cm}
\end{figure*}

\begin{figure*}[t]
    \centering

    \includegraphics[width=1.0\linewidth]{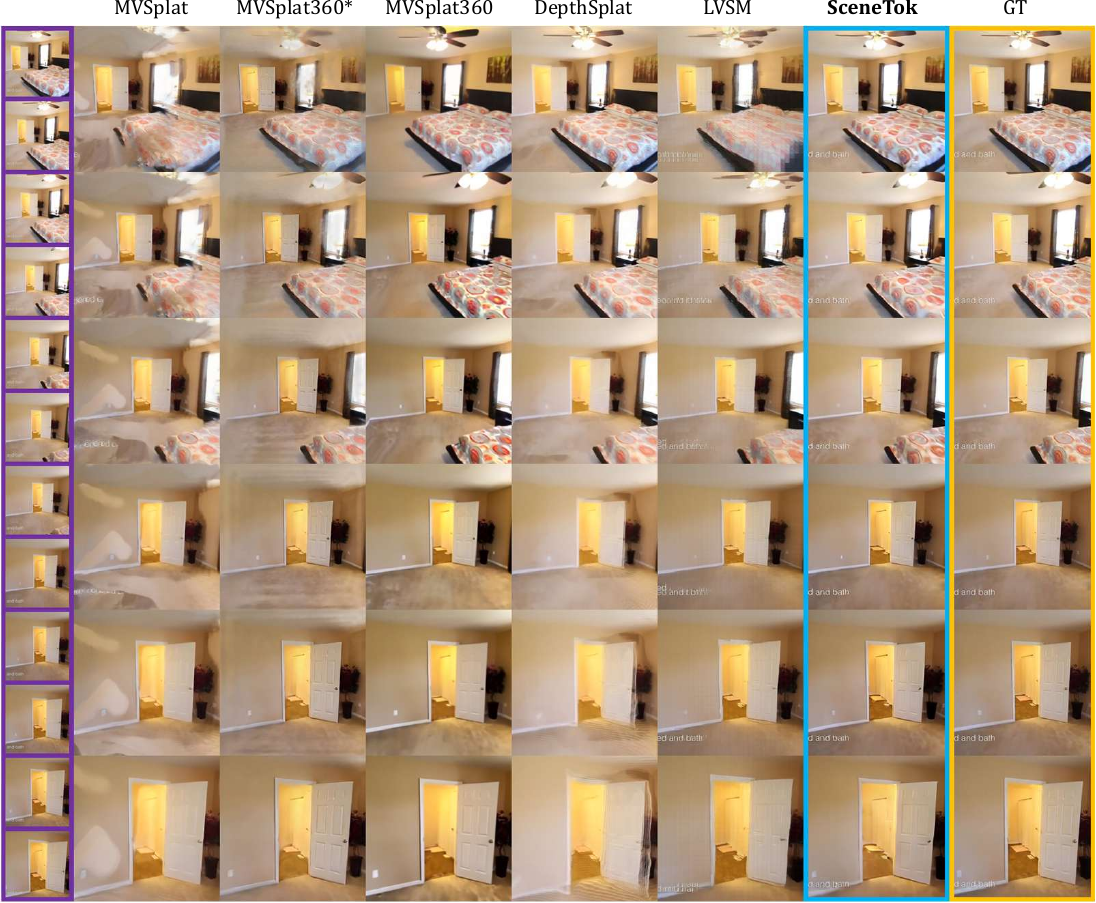}
    \vspace{-0.4cm}
    \caption{\textbf{Qualitative NVS on RealEstate10K.} $($\textcolor{purple}{\thickBox}, \textcolor{ours}{\thickBox}, \textcolor{orange}{\thickBox}$)$ denotes the 12 input context views, the target renderings from \ours and the corresponding ground-truth target views respectively (we show six views from top-to-bottom). * denotes output of MVSplat360 taken before the refinement step with a video diffusion model.
    }
    \label{fig:qualitative_nvs_2}
    \vspace{-0.4cm}
\end{figure*}

\begin{figure*}[t]
    \centering

    \includegraphics[width=1.0\linewidth]{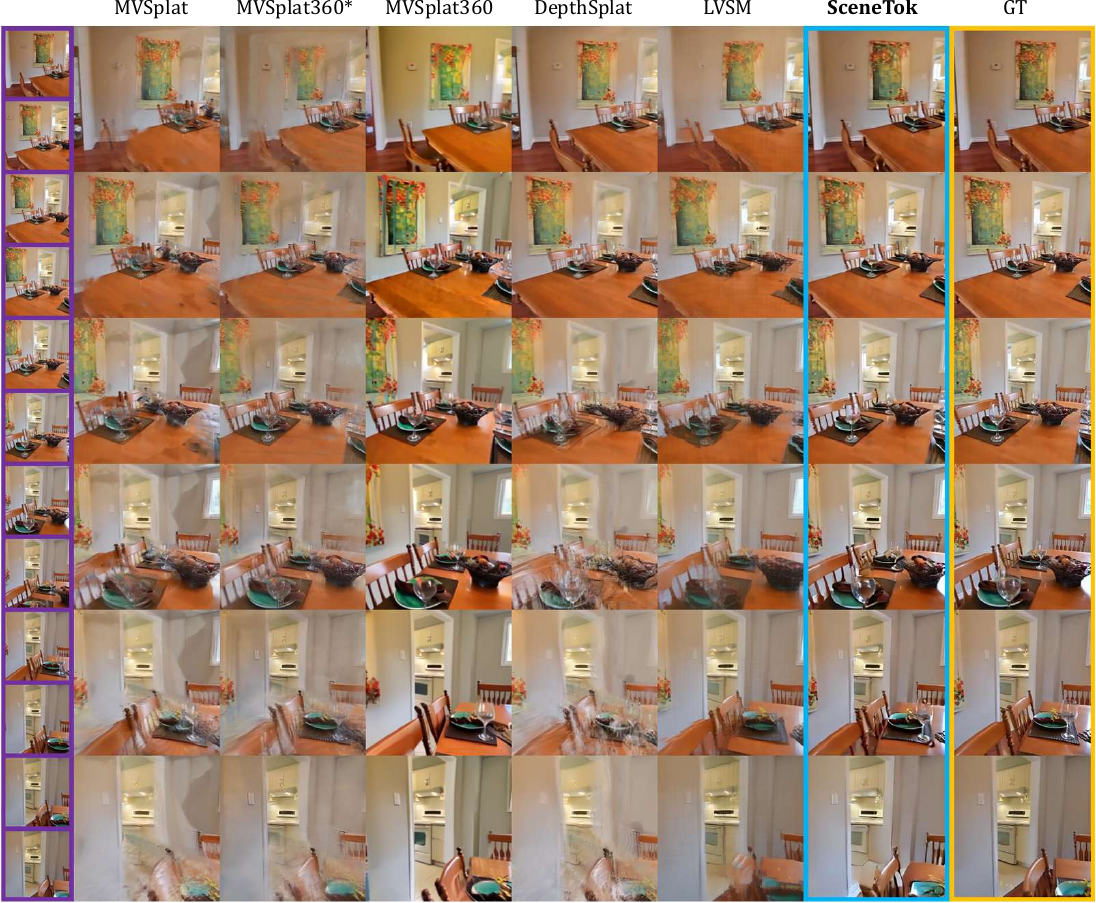}
    \vspace{-0.4cm}
    \caption{\textbf{Qualitative NVS on RealEstate10K.} $($\textcolor{purple}{\thickBox}, \textcolor{ours}{\thickBox}, \textcolor{orange}{\thickBox}$)$ denotes the 12 input context views, the target renderings from \ours and the corresponding ground-truth target views respectively (we show six views from top-to-bottom). * denotes output of MVSplat360 taken before the refinement step with a video diffusion model.
    }
    \label{fig:qualitative_nvs_3}
    \vspace{-0.4cm}
\end{figure*}

\begin{figure*}[t]
    \centering

    \includegraphics[width=1.0\linewidth]{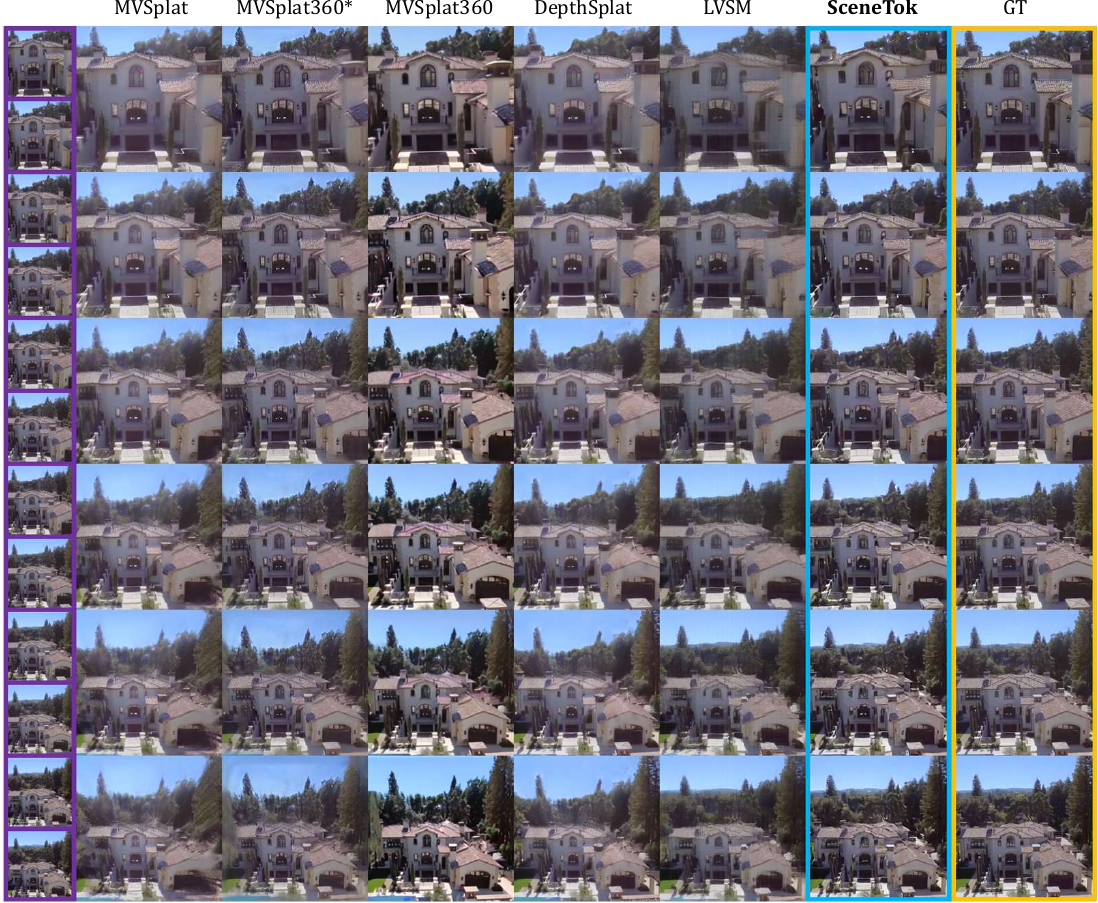}
    \vspace{-0.4cm}
    \caption{\textbf{Qualitative NVS on RealEstate10K.} $($\textcolor{purple}{\thickBox}, \textcolor{ours}{\thickBox}, \textcolor{orange}{\thickBox}$)$ denotes the 12 input context views, the target renderings from \ours and the corresponding ground-truth target views respectively (we show six views from top-to-bottom). * denotes output of MVSplat360 taken before the refinement step with a video diffusion model.
    }
    \label{fig:qualitative_nvs_4}
    \vspace{-0.4cm}
\end{figure*}

\begin{figure*}[t]
    \centering

    \includegraphics[width=1.0\linewidth]{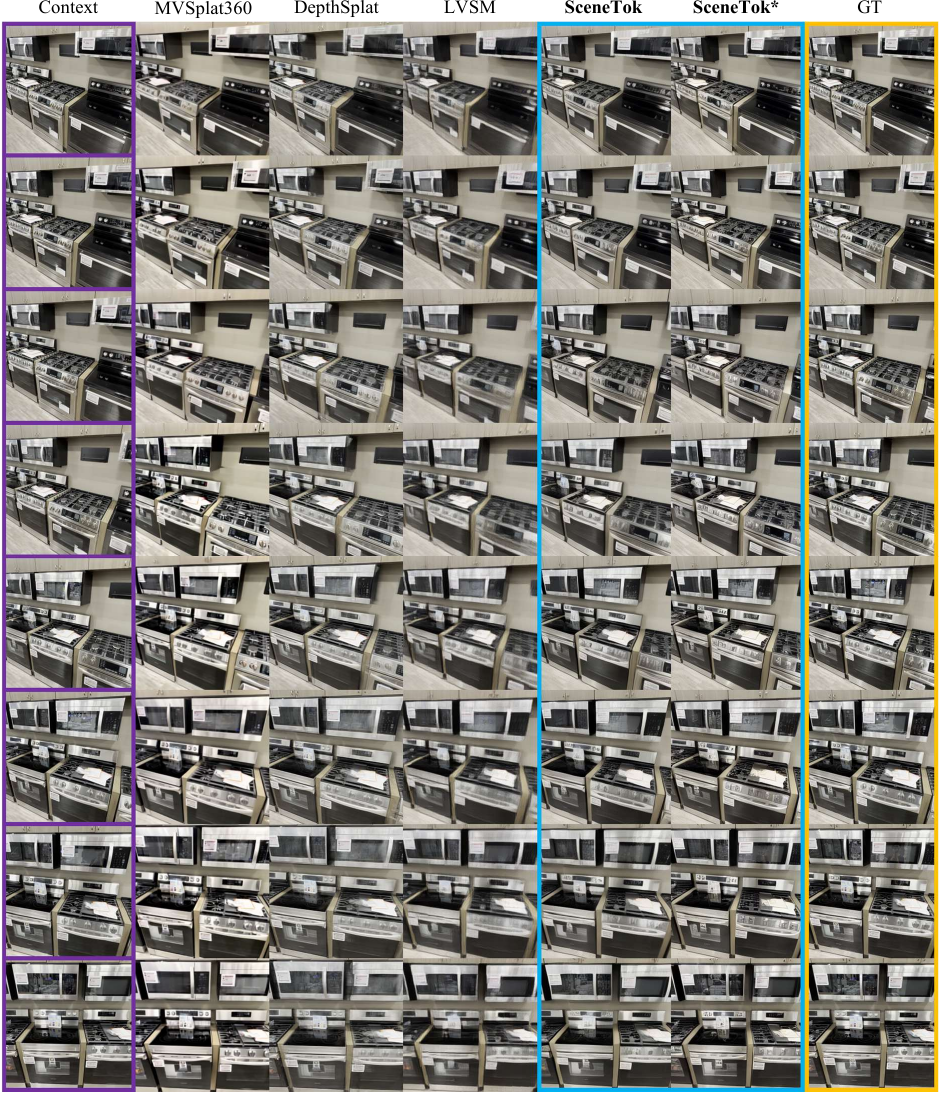}
    \vspace{-0.4cm}
    \caption{\textbf{Qualitative NVS on DL3DV-140.} $($\textcolor{purple}{\thickBox}, \textcolor{ours}{\thickBox}, \textcolor{orange}{\thickBox}$)$ denotes the input context views (only 8 are shown), the target renderings from \ours (* denote decoder version trained with WanVAE~\cite{wan2025}) and the corresponding ground-truth target views respectively (we show eight views from top-to-bottom). 
    }
    \label{fig:qualitative_nvs_5}
    \vspace{-0.4cm}
\end{figure*}

\begin{figure*}[t]
    \centering

    \includegraphics[width=1.0\linewidth]{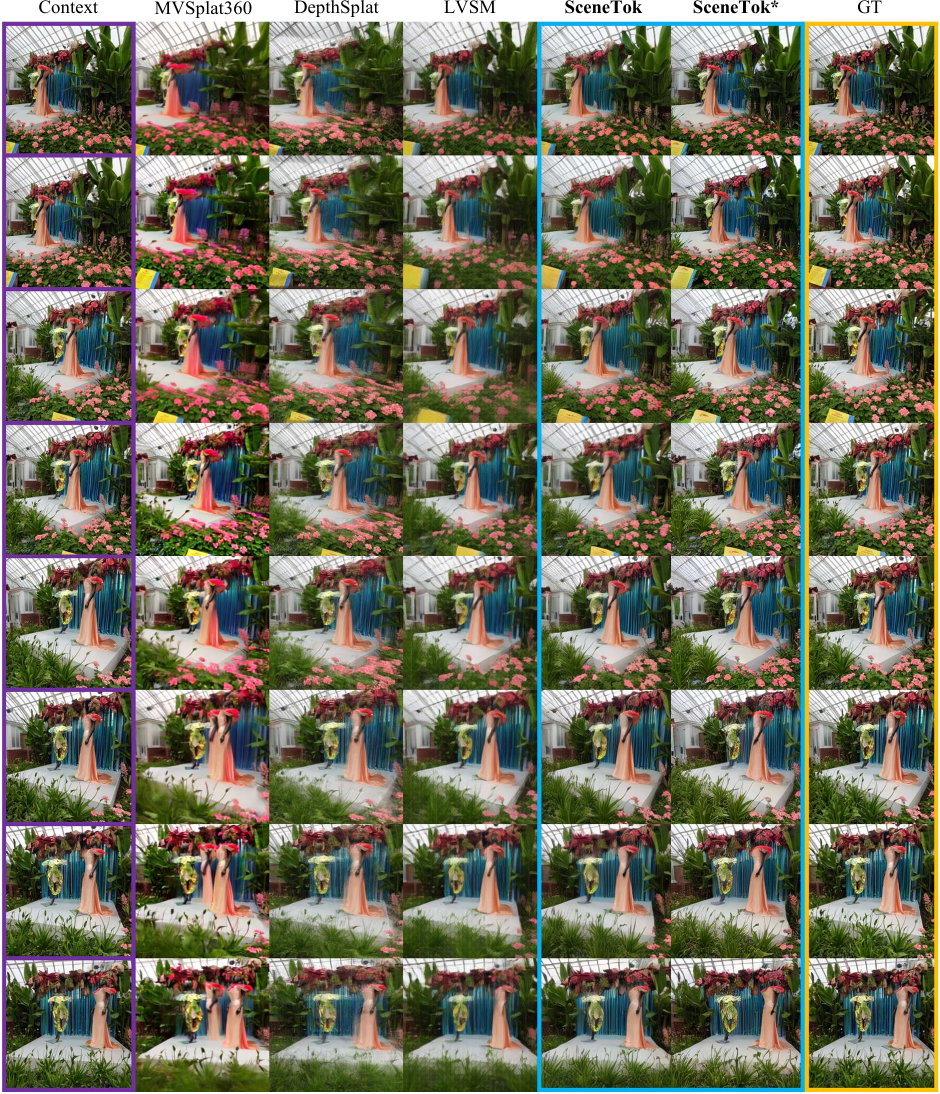}
    \vspace{-0.4cm}
    \caption{\textbf{Qualitative NVS on DL3DV-140.} $($\textcolor{purple}{\thickBox}, \textcolor{ours}{\thickBox}, \textcolor{orange}{\thickBox}$)$ denotes the input context views (only 8 are shown), the target renderings from \ours (* denote decoder version trained with WanVAE~\cite{wan2025}) and the corresponding ground-truth target views respectively (we show eight views from top-to-bottom). 
    }
    \label{fig:qualitative_nvs_6}
    \vspace{-0.4cm}
\end{figure*}

\begin{figure*}[t]
    \centering

    \includegraphics[width=1.0\linewidth]{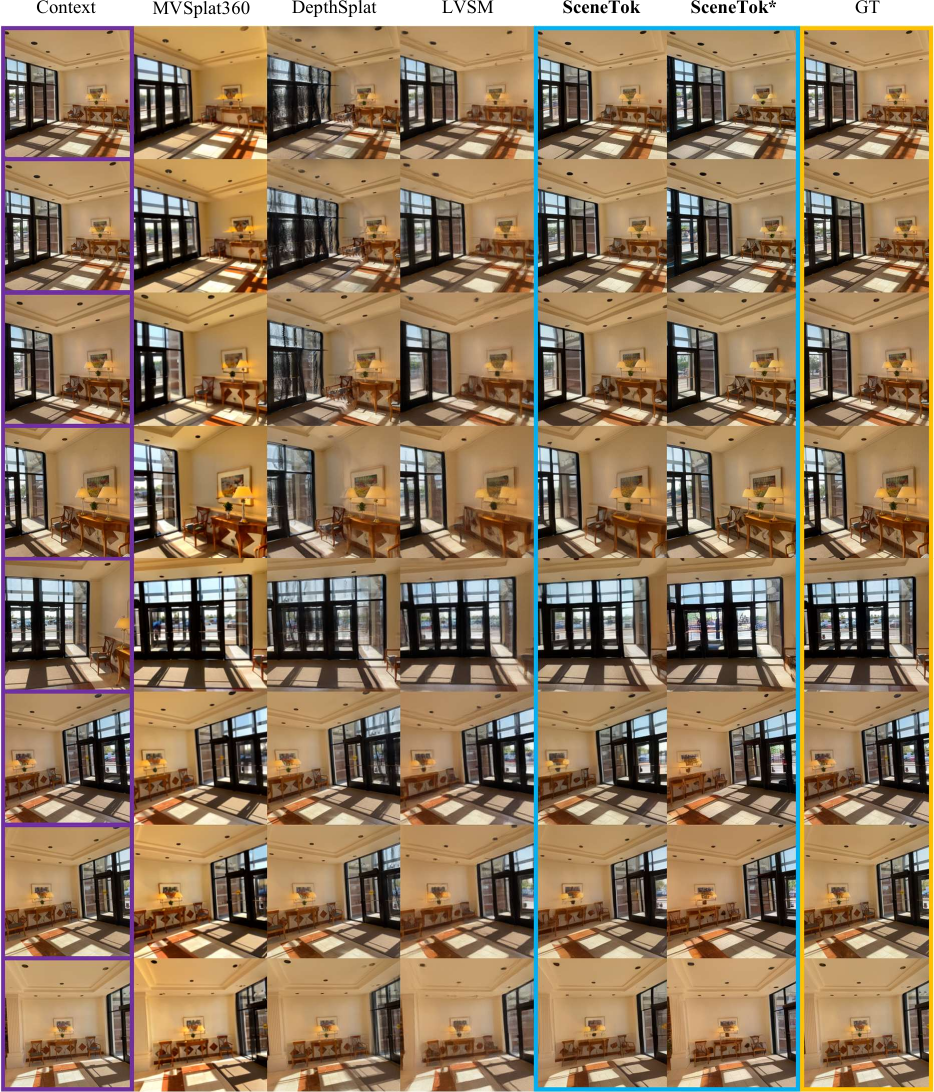}
    \vspace{-0.4cm}
    \caption{\textbf{Qualitative NVS on DL3DV-140.} $($\textcolor{purple}{\thickBox}, \textcolor{ours}{\thickBox}, \textcolor{orange}{\thickBox}$)$ denotes the input context views (only 8 are shown), the target renderings from \ours (* denote decoder version trained with WanVAE~\cite{wan2025}) and the corresponding ground-truth target views respectively (we show eight views from top-to-bottom). 
    }
    \label{fig:qualitative_nvs_7}
    \vspace{-0.4cm}
\end{figure*}

\begin{figure*}[t]
    \centering

    \includegraphics[width=1.0\linewidth]{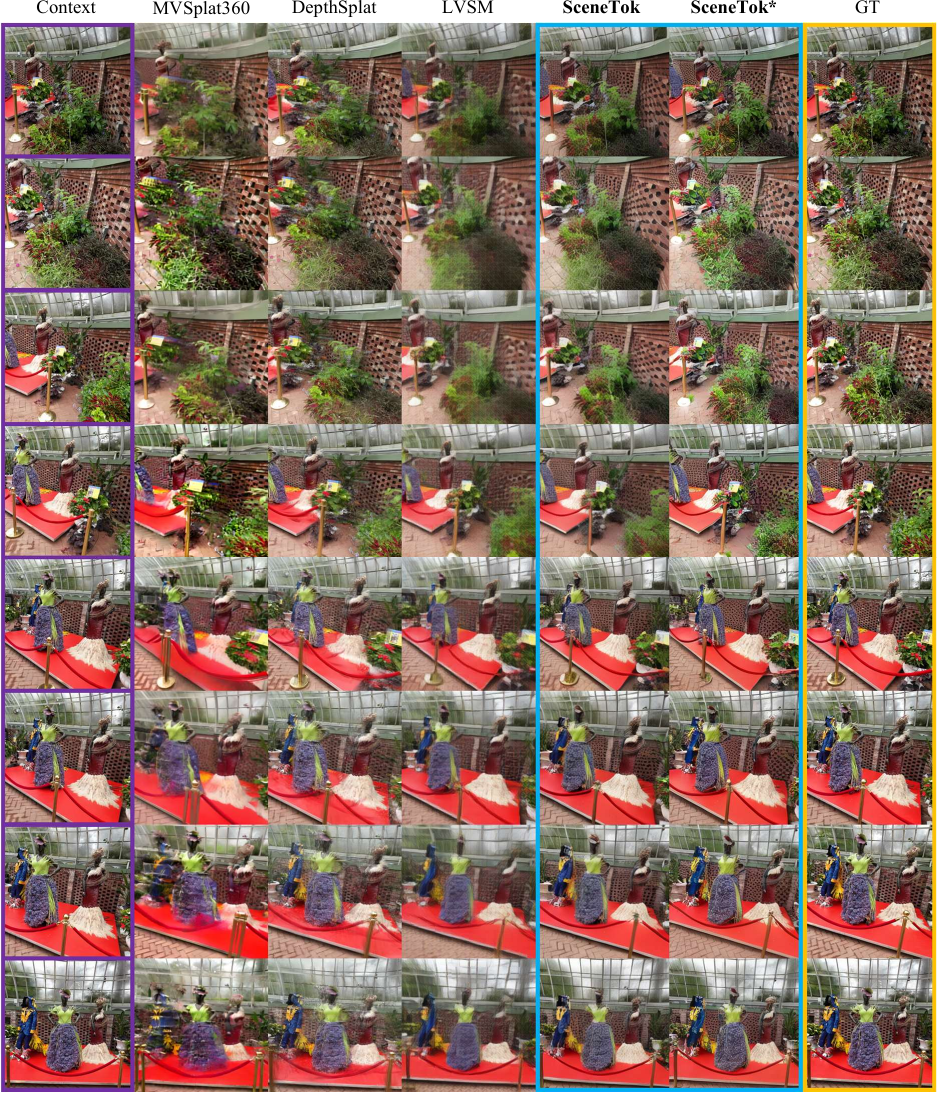}
    \vspace{-0.4cm}
    \caption{\textbf{Qualitative NVS on DL3DV-140.} $($\textcolor{purple}{\thickBox}, \textcolor{ours}{\thickBox}, \textcolor{orange}{\thickBox}$)$ denotes the input context views (only 8 are shown), the target renderings from \ours (* denote decoder version trained with WanVAE~\cite{wan2025}) and the corresponding ground-truth target views respectively (we show eight views from top-to-bottom). 
    }
    \label{fig:qualitative_nvs_9}
    \vspace{-0.4cm}
\end{figure*}

\begin{figure*}[t]
    \centering

    \includegraphics[width=1.0\linewidth]{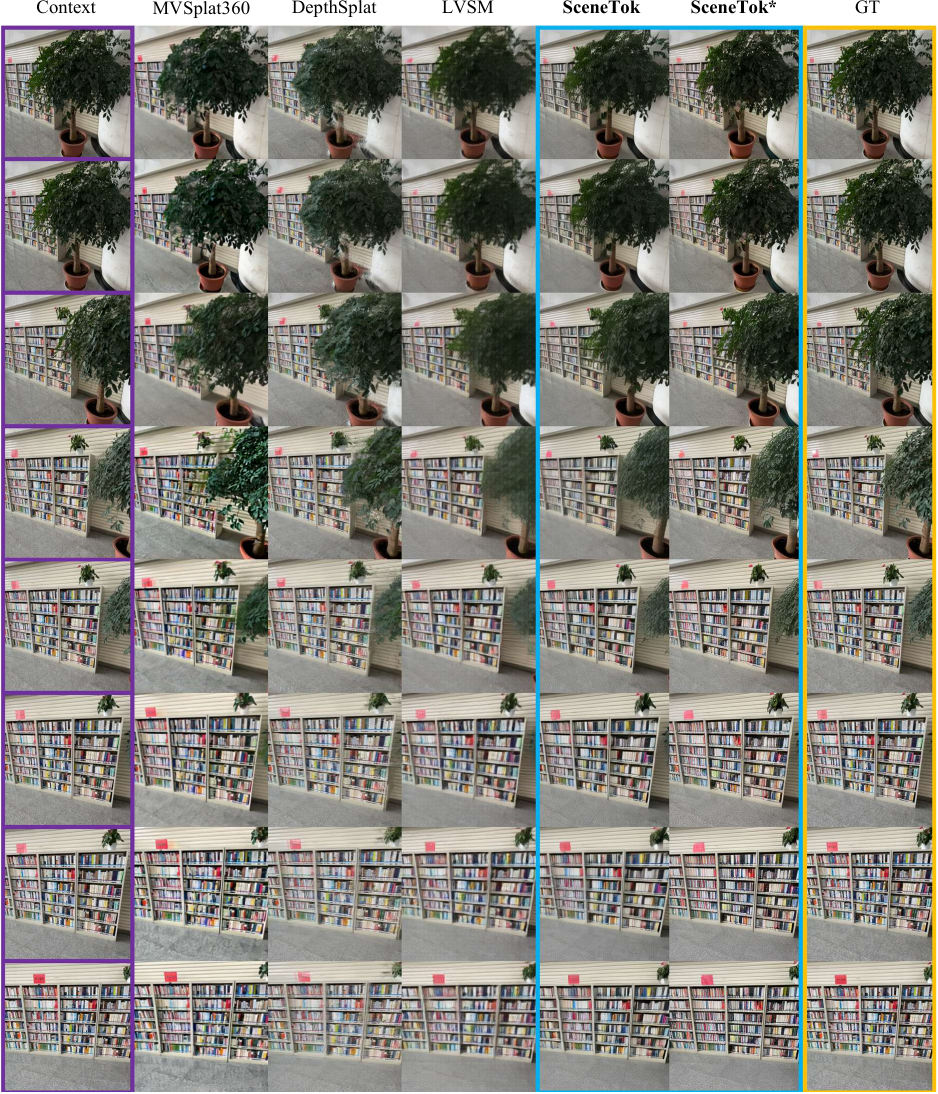}
    \vspace{-0.4cm}
    \caption{\textbf{Qualitative NVS on DL3DV-140.} $($\textcolor{purple}{\thickBox}, \textcolor{ours}{\thickBox}, \textcolor{orange}{\thickBox}$)$ denotes the input context views (only 8 are shown), the target renderings from \ours (* denote decoder version trained with WanVAE~\cite{wan2025}) and the corresponding ground-truth target views respectively (we show eight views from top-to-bottom). 
    }
    \label{fig:qualitative_nvs_10}
    \vspace{-0.4cm}
\end{figure*}

\begin{figure*}[t]
    \centering

    \includegraphics[width=1.0\linewidth]{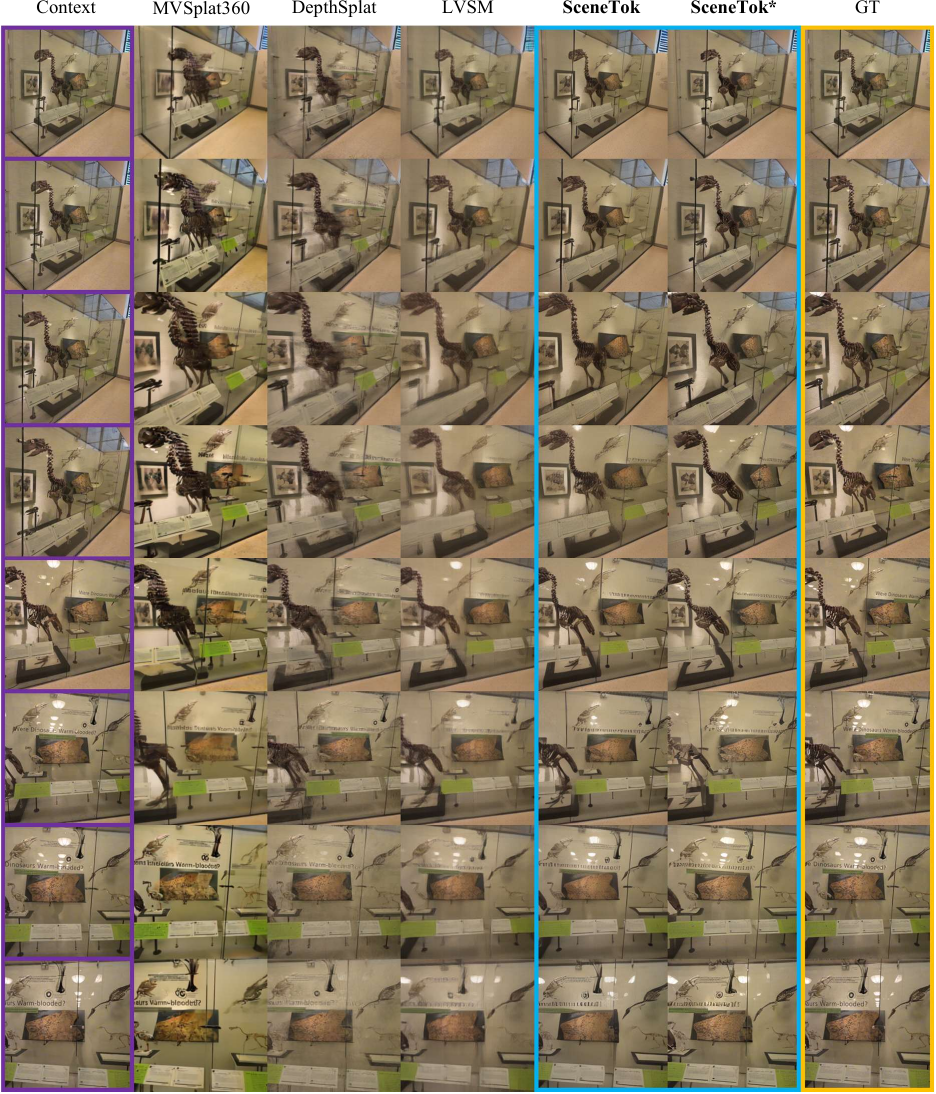}
    \vspace{-0.4cm}
    \caption{\textbf{Qualitative NVS on DL3DV-140.} $($\textcolor{purple}{\thickBox}, \textcolor{ours}{\thickBox}, \textcolor{orange}{\thickBox}$)$ denotes the input context views (only 8 are shown), the target renderings from \ours (* denote decoder version trained with WanVAE~\cite{wan2025}) and the corresponding ground-truth target views respectively (we show eight views from top-to-bottom). 
    }
    \label{fig:qualitative_nvs_11}
    \vspace{-0.4cm}
\end{figure*}

\paragraph{Novel-view synthesis.} We show qualitative examples for NVS in Fig.~\ref{fig:qualitative_nvs_1},~\ref{fig:qualitative_nvs_2},~\ref{fig:qualitative_nvs_3},~\ref{fig:qualitative_nvs_4},~\ref{fig:qualitative_nvs_5},~\ref{fig:qualitative_nvs_6},~\ref{fig:qualitative_nvs_7},~\ref{fig:qualitative_nvs_9},~\ref{fig:qualitative_nvs_10} and~\ref{fig:qualitative_nvs_11} on RealEstate10K~\cite{zhou2018stereo} and DL3DV~\cite{dl3dv}, and observe that \ours (both trained on VideoDCAE~\cite{opensora2} and WanVAE~\cite{wan2025}) achieves better high-frequency details, as indicated by the faithful reconstruction of the text within the image (visible in the first example), and is well represented by our method even under high compression. Overall, we observe better and cleaner renderings for our method in comparison to the existing works.

\begin{figure*}[t]
    \centering
    \begin{subfigure}[a]{1.0\linewidth}
        
        \includegraphics[width=1.0\linewidth]{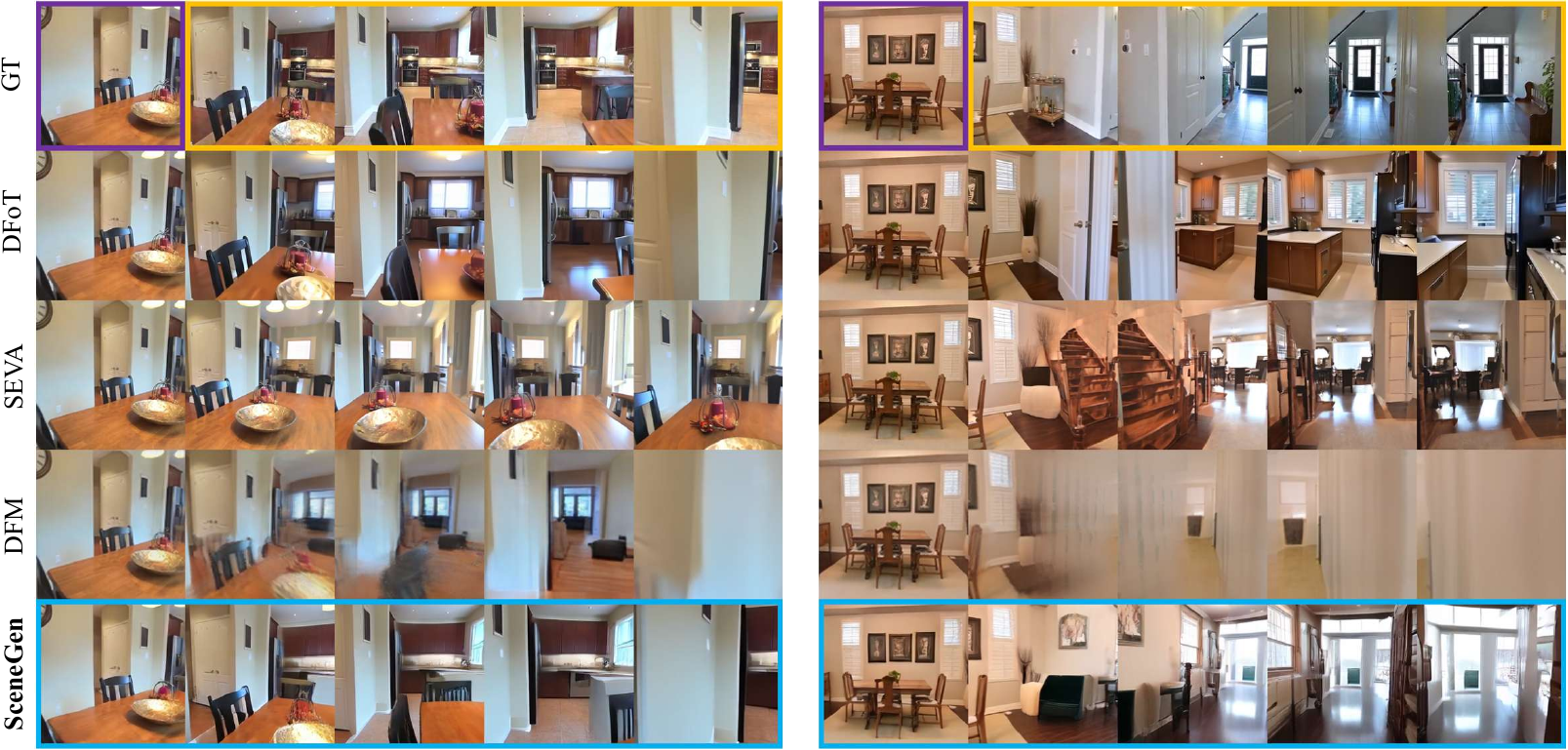}
        
    % \vspace{-0.2cm}
    \end{subfigure}
    \begin{subfigure}[b]{1.0\linewidth}
        
        \includegraphics[width=1.0\linewidth]{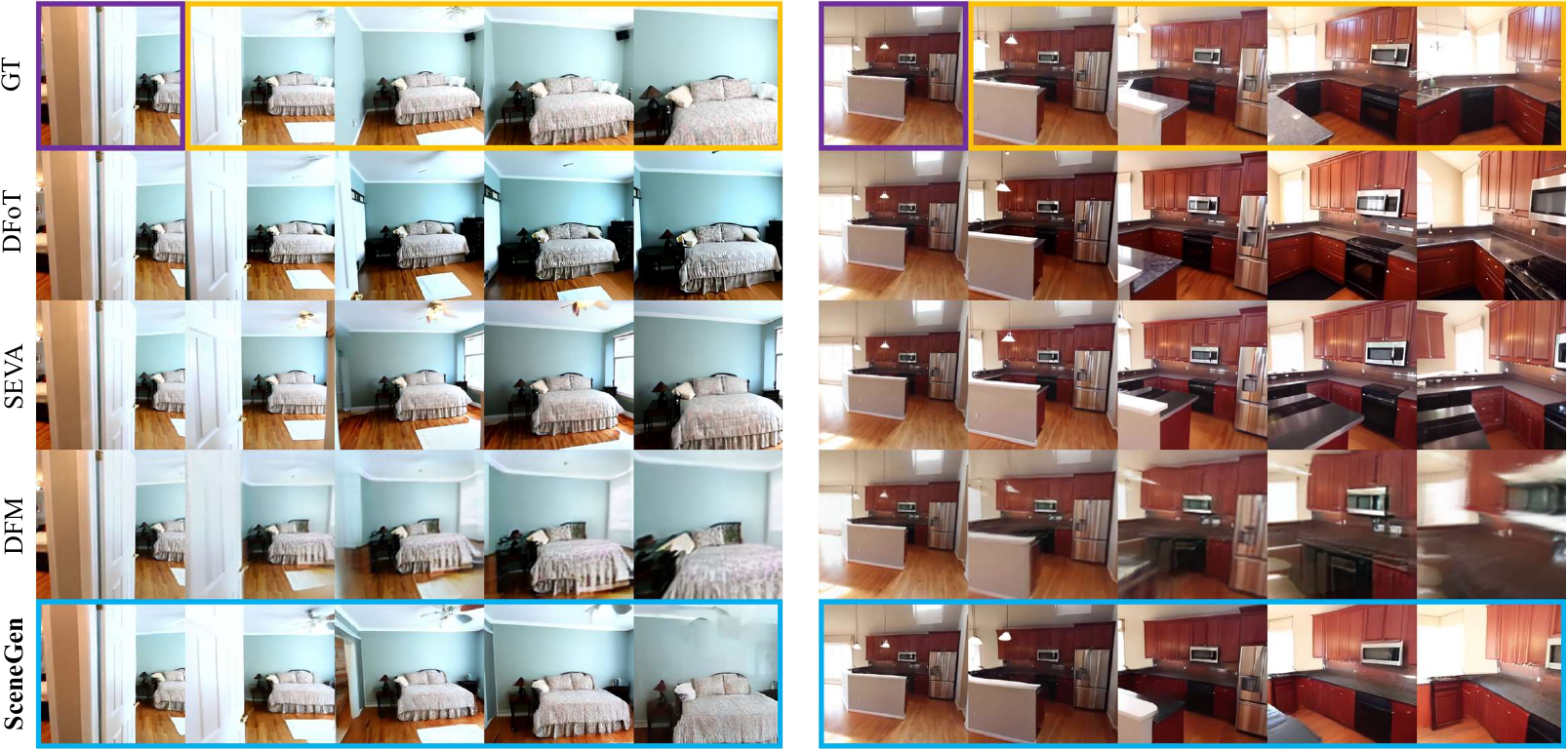}
        \vspace{-0.4cm}
    \end{subfigure}
    \caption{\textbf{Qualitative Single-View Generation Comparison.} $($\textcolor{purple}{\thickBox}, \textcolor{ours}{\thickBox}, \textcolor{orange}{\thickBox}$)$ denotes the input view, the target renderings from \ourscene and the corresponding ground-truth target views respectively.
    }
    \label{fig:qualitative_single_gen_2}
    \vspace{-0.4cm}
\end{figure*}

\begin{figure*}[t]
    \centering
    \begin{subfigure}[a]{1.0\linewidth}
        
        \includegraphics[width=1.0\linewidth]{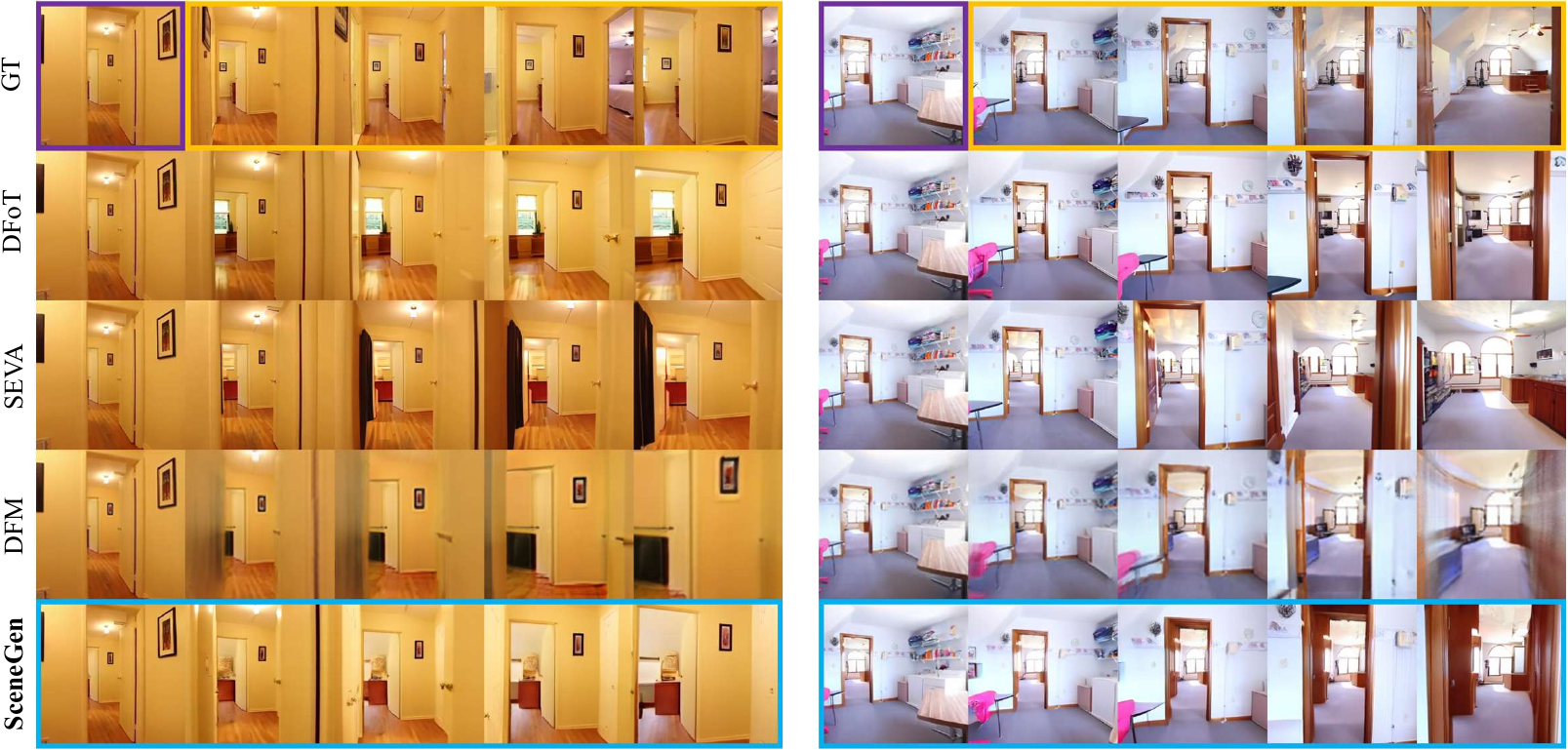}
        
    % \vspace{-0.2cm}
    \end{subfigure}
    \begin{subfigure}[b]{1.0\linewidth}
        
        \includegraphics[width=1.0\linewidth]{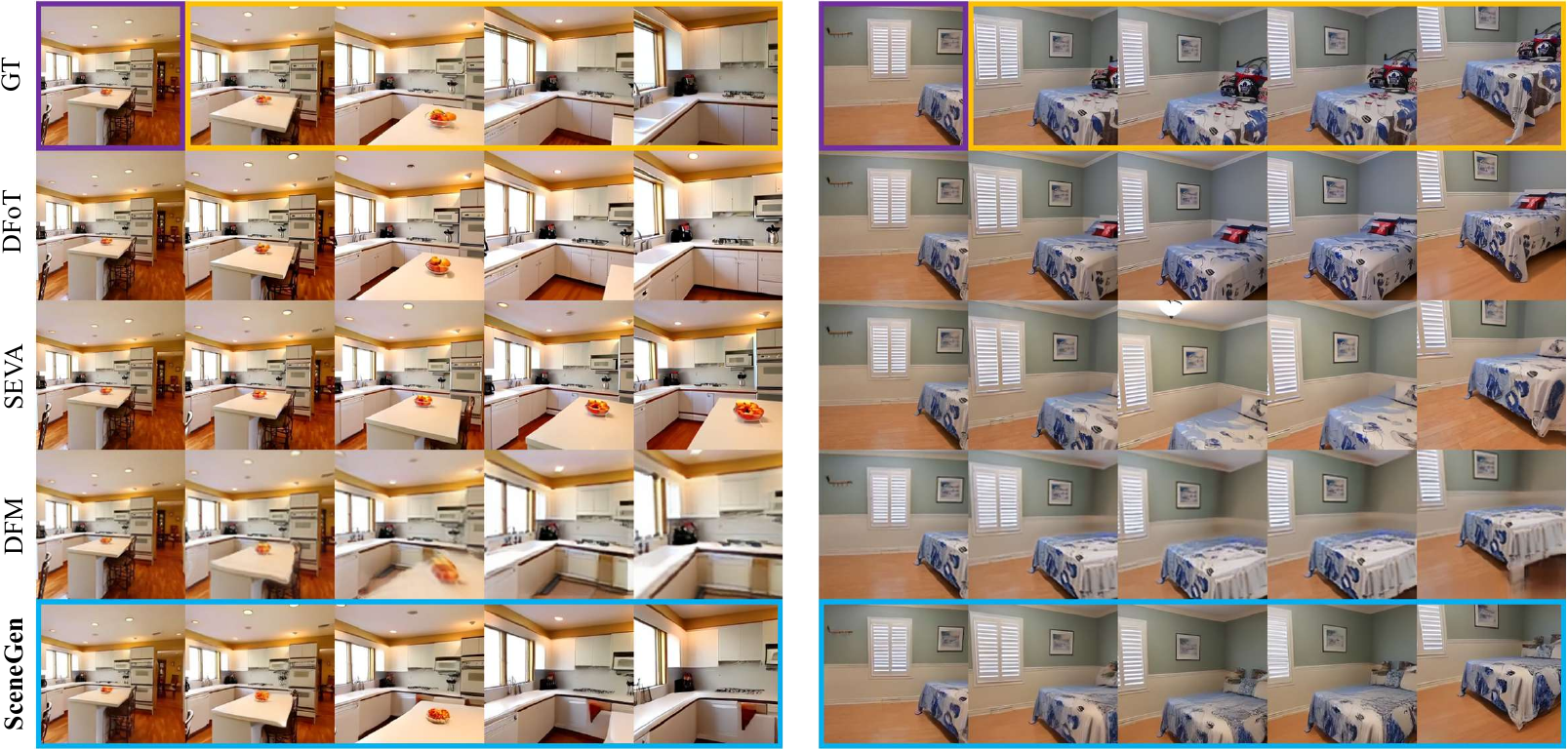}
        \vspace{-0.4cm}
    \end{subfigure}
    \caption{\textbf{Qualitative Single-View Generation Comparison.} $($\textcolor{purple}{\thickBox}, \textcolor{ours}{\thickBox}, \textcolor{orange}{\thickBox}$)$ denotes the input view, the target renderings from \ourscene and the corresponding ground-truth target views respectively.
    }
    \label{fig:qualitative_single_gen_3}
    \vspace{-0.4cm}
\end{figure*}

\vspace{-0.3cm}
\paragraph{Single-view generation.} We provide additional qualitative results for single-view generation in Fig.~\ref{fig:qualitative_single_gen_2}, and~\ref{fig:qualitative_single_gen_3}. Overall, we observe comparable generation quality, and our method is able to produce complete structures and geometry with partial observations. In comparison to DFoT~\cite{dfot} and SEVA~\cite{seva}, our method falls short of generating high-frequency details.



\end{appendices}

\end{document}